\documentclass{gCMB2e}

\usepackage{subfigure}% Support for small, `sub' figures and tables

\newtheorem{defn}{Definition}

\numberwithin{equation}{section}
\theoremstyle{plain}

\begin{document}

%\jvol{00} \jnum{00} \jyear{2014} \jmonth{February}

\articletype{RESEARCH ARTICLE}

\title{{\itshape Active skeleton for bacteria modeling}}

\author{Jean-Pascal Jacob$^{\rm a}$ and Mariella Dimiccoli$^{\rm b}$$^{\ast}$\thanks{$^\ast$Corresponding author. Email: mariella.dimiccoli@cvc.uab.es
\vspace{6pt}} and Lionel Moisan$^{\rm a}$\\\vspace{6pt} $^{a}${\em{Universit\'e Paris Descartes, MAP5 (CNRS UMR 8145), Paris, France}};
\vspace{6pt}$^{b}${\em{Computer Vision Center (CVC) and University of Barcelona (UB), Barcelona Perceptual Computing Lab (BCNPCL), Barcelona, Spain }}\\\received{v4.0 released February 2014} }

\maketitle

\begin{abstract}
The investigation of spatio-temporal dynamics of bacterial cells and their molecular components requires automated image analysis tools to track cell shape properties and molecular component locations inside the cells. In the study of bacteria aging, the molecular components of interest are protein aggregates accumulated near bacteria boundaries.
This particular location makes very ambiguous the correspondence between aggregates and cells, since computing accurately bacteria boundaries in phase-contrast time-lapse imaging is a challenging task. 
This paper proposes an active skeleton formulation for bacteria modeling which provides several advantages: an easy computation of shape properties (perimeter, length, thickness, orientation), an improved boundary accuracy in noisy images, and a natural bacteria-centered coordinate system that permits the intrinsic location of molecular components inside the cell. Starting from an initial skeleton estimate, the medial axis of the bacterium is obtained by minimizing an energy function which incorporates bacteria shape constraints. Experimental results on biological images and comparative evaluation of the performances validate the proposed approach for modeling cigar-shaped bacteria like {\it Escherichia coli}. The Image-J plugin of the proposed method can be found online at \textit{http://fluobactracker.inrialpes.fr}.


\end{abstract}

\section{Introduction}\label{introduction}
\begin{figure}[h]
\centering
\includegraphics[width=58mm]{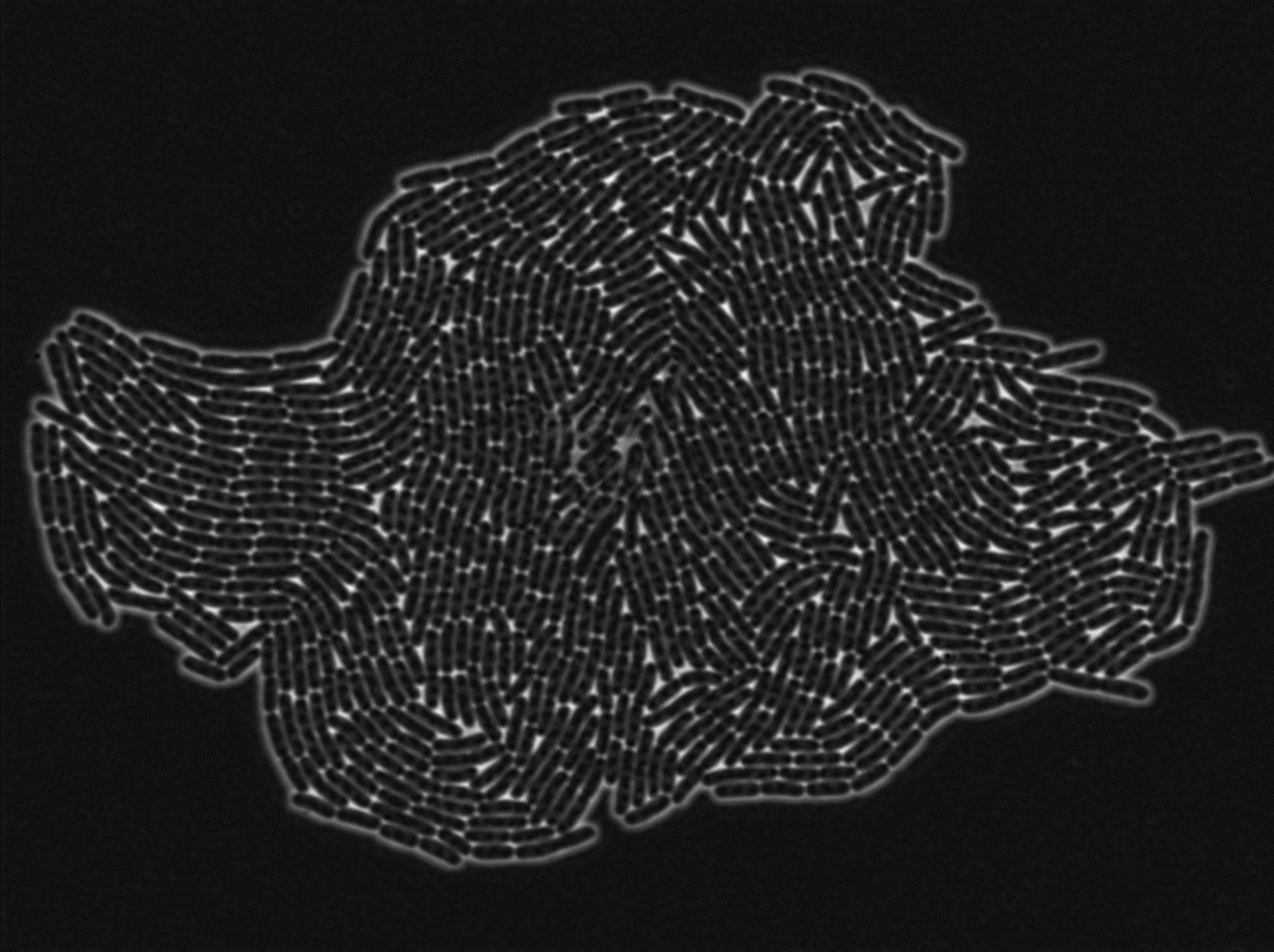}
\includegraphics[width=58mm]{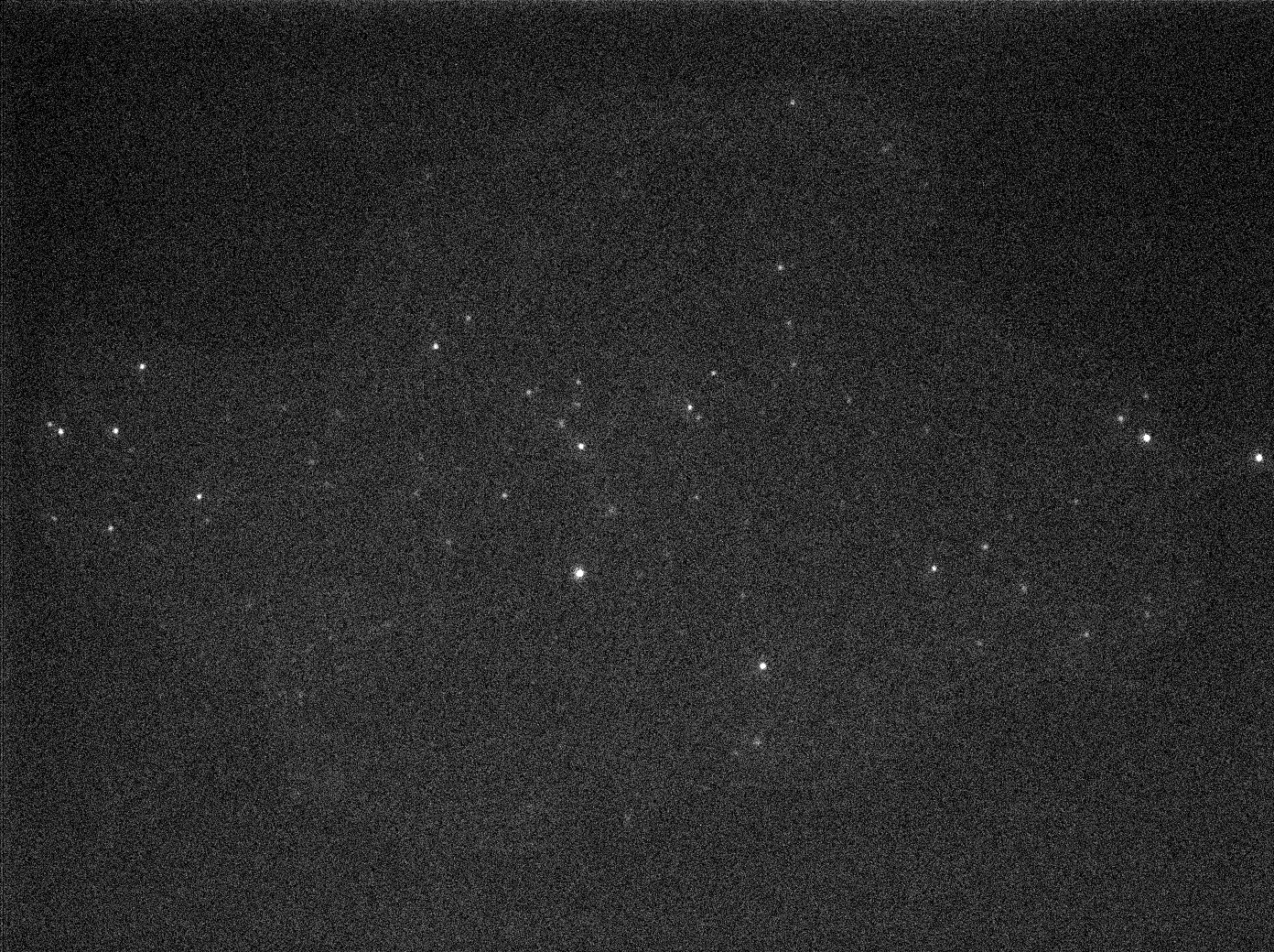}
\caption{{\it E. coli} cells visualized through phase-contrast microscopy (left), and their molecular components (in this case proteine aggregates) visualized through fluorescence microscopy (right). The cell size is approximately between 3 and 5 $\mu m$.}\label{example}
\end{figure}

\begin{figure}[h]
\centering
\includegraphics[width=50mm]{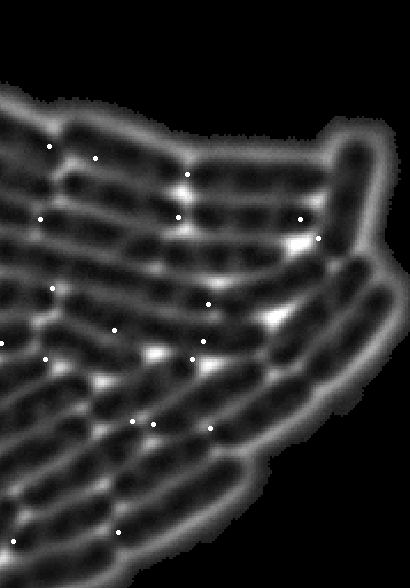}
\includegraphics[width=50mm]{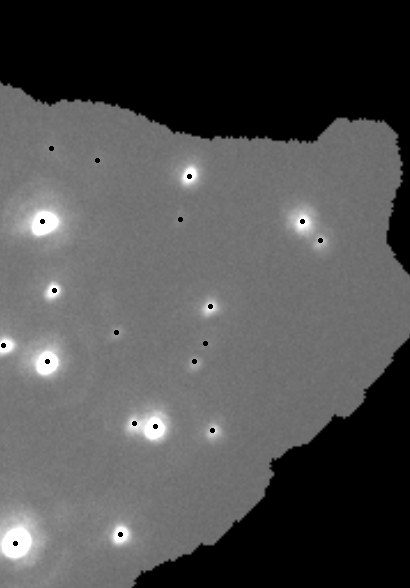}
\caption{{\it E. coli} cells visualized through phase-contrast microscopy (left) and protein aggregates visualized through a fluorescence microscopy (right). For a better visualization, here it is shown a saturated version of the original fluorescence image. Protein aggregates detected by \cite{dimiccoli2015particle} on the original fluorescence image,  are visualized as white points on left image. As it can be seen, they are often localized near boundaries making ambiguous the assignment to the host cell. The cell size is approximately between 3 and 5 $\mu m$.}\label{example_ambiguity}
\end{figure}

One of the fundamental issues addressed by Computational Cell Biology is the characterization of spatio-temporal dynamics of bacterial cells and their molecular components \citep{Slepchenko2002Computational}.
The rapid development of techniques for fluorescence imaging in recent years has opened new research opportunities, 
that need to cope with the availability of automated image analysis tools to be fully exploited. In the study of \textit{E. Coli} aging \citep{lindner2008asymmetric}, the molecular components of interest are protein aggregates accumulated at the old pole region of the aging bacterium. These subcellular components, visualized by fluorescence imaging techniques, need to be reliably associated to their host cells, which are generally visualized with phase-contrast microscopy. Figure~\ref{example} shows a pair of simultaneous frames extracted from a time-lapse phase-contrast and fluorescence image sequence. The automatic association between aggregates and bacteria requires a very accurate estimation of bacteria boundaries in this case, because aggregates accumulate near these boundaries (see Figure~\ref{example_ambiguity}).
In addition to boundary accuracy, a requirement for bacteria modeling is the possibility to derive accurate cell measurements such as length, width, orientation, perimeter, etc., that are fundamental characteristics of bacteria \citep{Osborn2007Cell}. For instance, in bacteria like \textit{E. Coli}, tracking population variability over time helps to understand the combinations of effects of genetic and environmental factors on cell phenotype.
Finally, having the possibility to describe the location of subcellular components in an intrinsic cell-centered coordinate system may be very useful to characterize intracellular dynamics \citep{Coquel2013Localization}.

\hspace{-5cm}

\hspace{-5cm}

This paper aims to provide an algorithm able to extract from an image a complete parametric description of cigar- or rod-shaped bacteria (e.g., \textit{E. coli}) that it contains. %, allowing the accurate computation of bacteria properties such as length, thickness, orientation, boundaries and providing a bacterium-centered reference system into which to express the coordinates of subcellular components. 
The underlying mathematical representation is a curvilinear skeleton (also called medial axis), which defines, after an appropriate (non-constant) dilation, the boundary of the bacterium. This model is built by evolving a first skeleton estimate so as to minimize an energy that promotes a good image fitting while enforcing the expected properties of the cell shape. 
In the following, we present the state of the art related to the use of skeletons for shape characterization and active contours, on which the proposed model relies.
\subsection{Background}
Segmentation, the task of partitioning an image into coherent %meaningful 
regions, is ill-posed and simple continuity or homogeneity assumptions cannot cope with the large variability of object appearances. In some specific applications for which one expects to see objects from a particular class, prior knowledge (e.g., clues on appearance or shape) may be exploited to reduce the space of variations.
%The appartence to a particular class of objects is generally expressed through cues that generalize across the class such as appearence or shape. 
Such cues have traditionally been used by the Computer Vision community for object recognition, but their usefulness in segmentation has only more recently reached a common consensus \citep{Leibe2006Implicit}. Appearance-based methods rely on features points as Scale-Invariant Feature Transform (SIFT) \citep{Lowe2001Local} but their efficiency has proved to be limited since in general, object identity is more a function of form that a function of appearance \citep{Siddiqi2001Psychophysics,Siddiqi1996Shock,Siddiqi1995Parts,Leordeanu2007beyondlocal}.  
Furthermore, in applications such as living cell imaging, where the intensities are reduced to a minimum to avoid photodamage and photobleaching, key features may not be sufficiently available.

In the use of shape as an alternative or to reinforce the use of appearance, the skeleton, defined by Blum \citep{Blum1973Biological} as the set of medial loci of the maximal disks contained inside the object, has received much attention in the literature \citep{Bai2008Path,Siddiqi1998Shock,Zhu1995FORMS}. Compared to other shape representation such as edge fragments and shock graphs, the skeleton offers more flexibility in modeling spatial relationships between parts and has a good repeatability/distinctiveness trade-off. Although its usefulness for shape matching and classification on silhouettes has been recognized since the nineties, it is only recently that skeletons have been exploited for non-rigid object detection \citep{Bai2009Active,Trinh2011Skeleton}. In all these methods, the object is first recognized using reference skeletons, and then the skeleton is used to optimize an initial segmentation.

%Bai et al. \cite{Bai2009Active} proposed a detection algorithm that first learn the skeleton configurations from training samples in a weakly supervised manner and then matchs previously extracted edges map against the parts of learned templates. The result of the matching is a coarse level detection which is refined by allowing the template to rotate to reduce the distance matching.
%In \cite{Trinh2011Skeleton} the authors used a generative shape model to generate shape instance for a category by determining shared skeletal structures for that category. Then, they gauge these instances based on the extent of image evidence supporting each category hypothesis.  Finally, a Partitioned Chamfer Matching measure is used to capture the support of image edges for an hypotized shape.

The idea of optimizing an initial segmentation, under some constraints, is at the core of active contour (also called {\it snakes}) methods, which specifically refine an initial approximate contour according to the image data. %instead of an initial skeleton. 
Early active contour models~\citep{kass88,osher88,cohen93,caselles93} act by minimizing an energy function consisting of a data term (including image information), and a regularization term which, in general, imposes smoothness constraints on the object shape. 
According to the nature of the image information included in the data term, the existing active contour methods can be categorized into two types: edge-based models and region-based models. Edge-based models use local image information ---typically, gradient information---~\citep{kass88,staib92,park01} to stop the contour evolution on the object boundary, while region-based approaches~\citep{cohen92,ivins95,zhu96,chesnaud99,chan01,jehan03,Zhang2010Active} use global image features %such as statistical information 
inside and outside the contour to control the evolution. Region-based methods relying on local information are able to segment images with non-homogeneous intensities \citep{Li2007Implicit, Li2008Minimization}. However, as detailed by Wang et al. \citep{Wang2009Active}, such localization property introduces many local minima of the nonconvex energy functional. Consequently, the result is more dependent on the initialization of the contour. The initialization issue itself is since long time the object of investigations \citep{cohen91,cohen93balloon,kass88,bajcsy89,jain98,cohen97,amini90using,geiger95} and, recently, it has been partially addressed by the formulation of convex energy functionals \citep{Mao2010Convex,Thieu2011Convex}, for which a single global minimum exists. According to the contour parametrization method used to model the smoothness constraint included in the regularization term, the existing active contour methods can be categorized into three classes: level sets snakes~\citep{osher88,caselles93,malladi95}, point-based snakes~\citep{kass88,xu98} and parametric snakes~\citep{staib92,brigger00}. Level sets define the 2D-curve implicitly from an evolving surface in a 3D space. They easily enable topology changes but are computationally expensive. The smoothness constraints are also implicit, that is, they are defined on the surface and not on the curve. Point-based snakes correspond to no parametrization or equivalently to a polygonal line (B-splines of degree zero). Parametric snakes define the curve between knot points \citep{brigger00,jacob04} via a basis decomposition such as B-splines~\citep{mark90,brigger00} or Fourier exponentials~\citep{staib92}. The overall smoothness is ensured with curvature and eventually length minimization. When a shape-prior is introduced into point-based or parametric snakes, a deformable template model is obtained~\citep{jain98}. The  prior shape model may be built ad hoc with analytical formulas~\citep{widrow73,yuille92,jolly96,lakshmanan96} or derived from a training set~\citep{staib92,cootes93,cootes95active,cootes95combining,davatzikos03}. For analytic models, the shape deformations are given by the model parameters (e.g., slope for straight lines, radius for circles). Mostly no probability distribution is needed. Training methods, usually called active shape models ~\citep{staib92, cootes95combining,davatzikos03}, are more flexible since they can be applied easily to any reproducible shape. However, they require to define an average shape and deformation models, which could bias the result if, for example, the training set is too small.

%~\cite{chakraborty96}

%Recently Mao et al. \cite{Mao2010Convex} proposed a convex neighbor-constrained active contour model to segment images with intensity non-homogeneity. With different shapes and sizes of the neighborhood for each point, theur model can capture the region information of a given image.
%Thieu et al \cite{Thieu2011Convex} proposed a novel model for active contours which is convex and therefore independent of the initial condition. Furthermore, the energy function of the proposed model is minimized in a computationally efficient way by using the Chambolle method.

%with selective local or global segmentationoWe don't focus on this since our initialization is close enough to the goal.
\subsection{Active skeleton}

Instead of matching a given bacteria template to a previously computed edge map as in classical recognition problems, or evolving an initial contour toward the object outline as in classical active-contour methods, we propose an active skeleton model which evolves initial skeletal polygonal lines toward the true medial axes of the bacteria. The advantage of the proposed approach for segmentation is that it allows to introduce strong shape constraints adapted to the bacteria class (here, cigar-shaped bacteria), which improves the accuracy of the boundary location even in very noised images. % that lead to accurate boundary location. 

% Then the resampling step common to classical active contour schemes is no more a very important issue. 

The paper is organized as follows. In Section~\ref{skel}, we describe the skeletal model we are considering, while
%and set the model definitions and the initialization. 
Section~\ref{snake} details the active skeleton initiation and the evolution process based on a variational (energy minimization) formulation. 
Section~\ref{result} provides results and comparisons to other segmentation methods.

\section{Method} \label{skel}

\subsection{Skeleton model}
We model a digital image acquired by a contrast-phase microscopy as a real valued discrete function $u:\Omega \subset \mathcal{Z}^2\rightarrow \mathbf{R}$, where $\Omega = [0,N]\times[0,M]$ is a rectangle.
%$\mathcal{S}$ is a kind of simplified skeleton described by $n$ triplets. %since each point $y$ belonging to a segment $[x_i,x_{i+1}]$ can be written as $y = (1-\lambda) x_i + \lambda x_{i+1}$, with $\lambda \in [0,1]$, its associated distance from the boundary can be computed accordly as  $r = (1-\ro) r_i + \ro r_{i+1}$. This is done by an orthognal dilation of the skeleton defined as follows. 
%The \textit{skeleton} $\mathcal{S} \in \Omega$ of a bacterial shape $\mathcal{S}$ on the image $u$ is the set of the medial loci of all maximal disks in $\mathcal{B}$.

\begin{defn}
A ball $B(x,r)$ centered at $x$ and having radius $r$ is a maximal disk of a shape $\mathcal{F}$ if $B(x,r)\subset \mathcal{F}$ and $\forall \tilde{B}(x',r')\subset \mathcal{F}, \tilde{B}(x',r') \neq B(x,r) \Rightarrow B(x,r)\not\subset \tilde{B}(x',r')$
\end{defn}
We denote the maximal disks of a shape by the symbol $B_m(x,r)$.
\begin{defn}
The morphological \textit{skeleton} $\mathcal{S} \in \Omega$ of a bacterium $\mathcal{B}$ on the image $u$ is the set of  maximal disks centers of $\mathcal{B}:  \mathcal{S}= \lbrace x | \exists r >0 : B(x,r) \equiv  B_m(x,r)\rbrace$.
\end{defn}

We represent a skeleton $\mathcal{S}$ by an ordered set of $n$ points with associated radius value: $\mathcal{S}=\{ (\vec{x_i},r_i)\}_{i\in \{1..n\}}$, where $\vec{x_i}=(x_i,y_i) \in \Omega$ represents the center of the maximal disk of radius $r_i$.  The skeleton is hence composed of $n-1$ segments $s_j = [\vec{x_j},\vec{x_{j+1}}]$, with $j \in [1,...,n-1]$. Given the skeleton $\mathcal{S}$, the bacterium is built up by dilating the centers of the maximal disks according to their associated radius, which is linearly interpolated inside each segment. Let $s_j = [\vec{x_j},\vec{x_{j+1}}]$ a segment of the skeleton, then for each point $\vec{x}_k$ of $s_j$, the corresponding radius is computed as a linear interpolation of $r_j$ and $r_{j+1}$: $r_k(\lambda)= (1- \lambda)r_j + \lambda r_{j+1}$, with $ 0 \leq \lambda \leq 1$.

Given an initial skeleton, the optimization of its location strongly depends on the underlying segment dilations and in turn to the definition of distance used for the dilation. %In the following, we introduce two different dilation models: a simplified one and the real one. The simplified one is a dilation orthogonal to the segments. It is directly linked to the usual distance definition $d_1$ which implies simpler derivatives calculus. Then it  has the advantage to be computationally faster. In our application to the modeling of Escherichia coli, the radius values vary slowly, so that the two models are almost equivalent. The skeleton dilation defined above, strictly depends on the metric used as distance. 
In the following, we define two distance models and, according to them,  we derive the expression of the dilation models.
\subsection{Distance models}
In this section, we  first define a simplified model of distance of a point to a segment, say $d_e$, and then we define an accurate orientation based distance, say $d_o$, that takes the radii difference into account.

\begin{figure}[h]
\centering
\includegraphics[width=7.3cm]{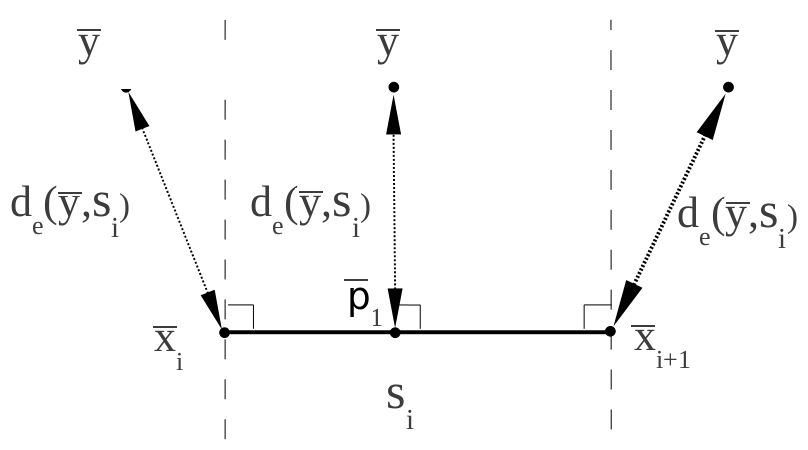} \\
\vspace{0.2cm}
\includegraphics[width=7.5cm]{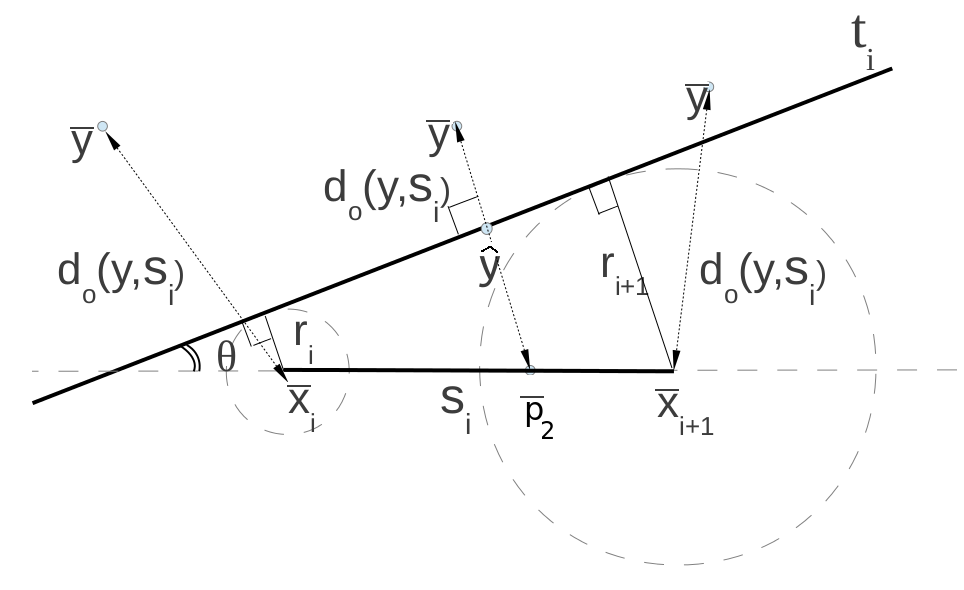} 
\vspace{-0.4cm}
\caption{Euclidean distance $d_e(y,s_i)$ (up) and orientation-based distance $d_o(y,s_i)$ (down) of a point $y$ to a skeletal segment $s_i$.}
\label{distances}
\end{figure}

\begin{defn}
\label{def2}
\textbf{Simplified distance of a point to a  segment.}  
Let $\vec{y}=(x,y) \in \Omega$ be any point of the image and $s_i=[\vec{x}_i, \vec{x}_{i+1}] \in \mathcal{S}$ any segment of the skeleton. Denoting by $\vec{p}_1$ the projection of $\vec{y}$ on the line through $\vec{x}_i$ and $\vec{x}_{i+1}$, $\vec{p}_1$ can be written as $\vec{p}_1 = (1-\lambda_1)\vec{x}_i + \lambda_1 \vec{x}_{i+1}$, with $\lambda_1 \in \mathbf{R}$.  The simplified Euclidean-based  distance of a point $\vec{y}$ to a segment $s_i$ is defined as follows.
\end{defn}

\begin{equation}
d_e(\vec{y},s_i)=
\left\{\begin{array}{lllcc}
||\vec{y}-\vec{x}_i|| \quad \lambda_1<0 \\
||\vec{y}-\vec{x}_{i+1}|| \quad \lambda_1>1 \\
||\vec{y}-\vec{p}_1|| \quad \lambda_1 \in ]0,1[ 
\label{eq:d_e}
\end{array}\right.
\end{equation}

where $|| \cdot ||$ is the Euclidean norm.
In Fig.\ref{distances} (up) it can be observed that the distance coincides with the Euclidean distance between two points only if the orthogonal projection of $\vec{y}$ on the line through $\vec{x}_i$ and $\vec{x}_{i+1}$ lies on the segment $s_i$.

The value of $\lambda_1$ can be computed from the scalar product $<\vec{y}-\vec{x}_i,\vec{x}_{i+1}-\vec{x}_i>$: $\lambda_1 = \frac{(x-x_i)\Delta x_i + (y-y_i)\Delta y_i}{L_i^2}$, where $\Delta x_i = x_{i+1} - x_i$, $\Delta y_i = y_{i+1} - y_i$, and $L_i = \sqrt{{\Delta x_i}^2 + {\Delta y_i}^2}$ (the length of segment $s_i$). 
If the orthogonal projection $\vec{p}_1$ of $\vec{y}$ on $s_i$ lies on $s_i$, then the distance of $\vec{y}$ to $s_i$ coincides to the distance of  $\vec{y}$ to $\vec{p}_1$. In this case, such a distance can be written as $d(y,s_i) = \sqrt{(x-x_i - \lambda_1\Delta x_i)^2+(y-y_i - \lambda_1\Delta y_i)^2}$. By introducing $\lambda_e$:

\begin{equation}
\lambda_e = \left\lbrace
\begin{array}{lllcc}
0 \quad \quad \lambda_1 \leq 0 \\
1 \quad \quad \lambda_1 \geq 1 \\
\lambda_1=\frac{(x-x_i)\Delta x_i + (y-y_i)\Delta y_i}{L_i^2} \quad 0 < \lambda_1 < 1
\label{eq:lambda_e}
\end{array}\right.
\end{equation}

the simplified distance from any point to a segment can be written as: $d_e(\vec{x},s_i)= \sqrt{(x-x_i- \lambda_e\Delta x_i)^2 + (y-y_i- \lambda_e\Delta y_i)^2}$.

\begin{defn}\label{def3}
\textbf{Orientation-based distance of a point to a  segment.}  

Let $\vec{y} \in \Omega$ be any point of the image and $s_i=[\vec{x}_i, \vec{x}_{i+1}] \in \mathcal{S}$ any segment of the skeleton. Let $\Delta r_i= | r_{i+1}-r_i|$ be the  difference between the radii value associated to the extremities of the segment $s_i$ and $L_i$ the segment length.
\begin{itemize}
\item  \textbf{Case 1: $L_i>\Delta r_i$}: let $t_i$ be the common tangent to the circles with  centers $(\vec{x}_i, \vec{x}_{i+1})$ and radii  $(r_i, r_{i+1})$  respectively, lying on the same half-plane than $\vec{y}$ compared to $t_i$. Let $\hat{y}$ be the orthogonal projection of $\vec{y}$ over $t_i$, and $\vec{p}_2$ the intersection of the line through $\vec{x}_i$ and $\vec{x}_{i+1}$ with the line through $\vec{y}$ and $\hat{y}$. Writing $\vec{p}_2$ as  $\vec{p}_2 =(1-\lambda_2) \vec{x}_i + \lambda_2 \vec{x}_{i+1}$, with $\lambda_2 \in \mathbf{R}$, the orientation-based distance $d_o$ is defined as follows: (see figure \ref{distances} (down)):
$$
d_o(\vec{y},s_i)=
\left\{\begin{array}{l}
||\vec{y}-\vec{x}_i|| \hspace{4mm} \lambda_2<0 \\

||\vec{y}-\vec{x}_{i+1}|| \hspace{4mm} \lambda_2>1 \\
||\vec{y}-\vec{p}_2|| \hspace{4mm} \lambda_2 \in ]0,1[ 
\end{array}\right.
$$
where $|| \cdot ||$ is the euclidean norm.
\item \textbf{Case 2: $L_i\leq\Delta r_i$}:
$$
d_o(\vec{y},s_i)=
\left\{\begin{array}{l}
||\vec{y}-\vec{x}_i||\hspace{4mm} r_i>r_{i+1}\\
||\vec{y}-\vec{x}_{i+1}|| \hspace{4mm} r_i<r_{i+1}
\end{array}\right.
$$
\end{itemize}
\end{defn}
 
%The skeleton's definition, that is the set of medial loci of the maximal disks inside object boundary, relis implicitly on the concept of distance of a point to a segment. 

Under the assumption $L_i>\Delta r_i$,  $\vec{p}_2= (1-\lambda_2)\vec{x}_i + \lambda_2 \vec{x}_{i+1}$ with $\lambda_2 \in \mathbf{R}$. Let $\lambda$ such that $\vec{p}_2-\vec{p}_1=\lambda(\vec{x}_{i+1}-\vec{x}_i)$. Then one has simply to replace $\lambda_1$ with $\lambda_2=\lambda_1+\lambda$ in the equations derived for the simplified distance.

To explicit $\lambda$, one can compute the angle $\theta$ between $t_i$ and $s_i$ according to $sin(\theta)=\frac{\Delta r_i}{L_i}$. It comes that 
\begin{equation}
\lambda =\frac{d_e}{L_i\sqrt{L_i^2-\Delta r_i^2}}\Delta r_i.
\end{equation}
Note again that $\lambda$ is an algebraic value. To deal with the case $L_i\leq\Delta r_i$ one can extend the definition of $\lambda$, for example in the following manner:
\begin{equation}
\tilde{\lambda}=
\left\{\begin{array}{l}
1-\lambda_1 \hspace{4mm}  \Delta r_i\geq L_i,\\
-\lambda_1 \hspace{4mm}  -\Delta r_i\geq L_i,\\
\lambda=\frac{d_e}{L_i\sqrt{L_i^2-\Delta r_i^2}}\Delta r_i \hspace{4mm}  otherwise.
\end{array}\right.
\end{equation}
Now one can compute in any case $\lambda_2=\lambda_1+\tilde{\lambda}$. Because of the restrictions of $\lambda_2$, the definition can be extended as:
\begin{equation}
\lambda_o=
\left\{\begin{array}{l}
0\ \hspace{4mm} \lambda_2 \leq 0,\\
1\ \hspace{4mm} \lambda_2 \geq 1, \\
\lambda_2= \lambda_1 +\tilde{\lambda} \hspace{4mm}  \lambda_2 \in ]0,1[, 
\end{array}\right.
\label{eq:lambda_o}
\end{equation}
Finally, the general expression of $d_o$ is as follows:
\begin{equation}
d_o(\vec{x},s_i)=\sqrt{(x-x_i-\lambda_o \Delta x_i)^2+(y-y_i-\lambda_o \Delta y_i)^2}.
\end{equation}

\subsection{Dilation models}
Before introducing the dilations models derived from the above definitions of distance, we define the distance of a point to as skeleton.

\begin{defn}\label{th2}
\textbf{Distance of a point to a skeleton.}
Let $\mathcal{S}=\{(\vec{x}_i,r_i)\}_{i = 1,...,n} \in \Omega$ a skeleton on the image $u:\Omega \rightarrow \mathbf{R}$ and $y$ a point of $\Omega$.
$$d(\vec{y},\mathcal{S})=\displaystyle{\min_{i = 1,...,n-1}}\Big \lbrace d(\vec{y},s_i) \Big \rbrace$$
\end{defn}

\begin{defn}
\label{defD1}
\textbf{Dilation model derived from the simplified distance of a point to a  segment.}  
Let $\mathcal{S}=\{ (\vec{x}_i,ri)\}_{i\in \{1..n\}}$ a skeleton and $s_j=[\vec{x}_j,\vec{x}_{j+1}]$  a segment of $\mathcal{S}$. For any point $\vec{y} \in \Omega$, let  $s_i=\underset{j \in [1,...,n]}{argmin} \hspace{1mm} d_e(\vec{y},s_j)$ the segment that has minimal simplified distance from $\vec{y}$. The simplified dilation of $\mathcal{S}$ is given by the set of points such that their simplified distance to $s_i$ is inside the maximal disks with interpolated radius:
$\mathcal{D}(\mathcal{S}) =  \lbrace \vec{x} | d_e(\vec{x},s_i) \leq (1-\lambda_e)r_i + \lambda_e r_{i+1} \rbrace$
where $\lambda_e$ is defined in equation \ref{eq:lambda_e}
\end{defn}

\begin{defn}
\label{defD2}
\textbf{Dilation model derived from orientation-based distance of a point to a segment.}  
Let $\mathcal{S}=\{ (\vec{x}_i,ri)\}_{i\in \{1..n\}}$ a skeleton and $s_j=[\vec{x}_j,\vec{x}_{j+1}]$  the $j-{th}$ segment of $\mathcal{S}$. For any point $\vec{y} \in \Omega$, let  $s_i=\underset{j \in [1,...,n]}{argmin} \hspace{1mm} d_o(\vec{y},s_j)$. The orientation-based dilation of $\mathcal{S}$ is: 

$\mathcal{D}(\mathcal{S}) =  \lbrace \vec{y} | d_o(\vec{y},s_i) \leq (1-\lambda_o)r_i + \lambda_o r_{i+1} \rbrace$
where $\lambda_o$ is defined in equation \ref{eq:lambda_o}.
\end{defn}

\begin{figure}[h]
\centering
\includegraphics[width=6.7cm]{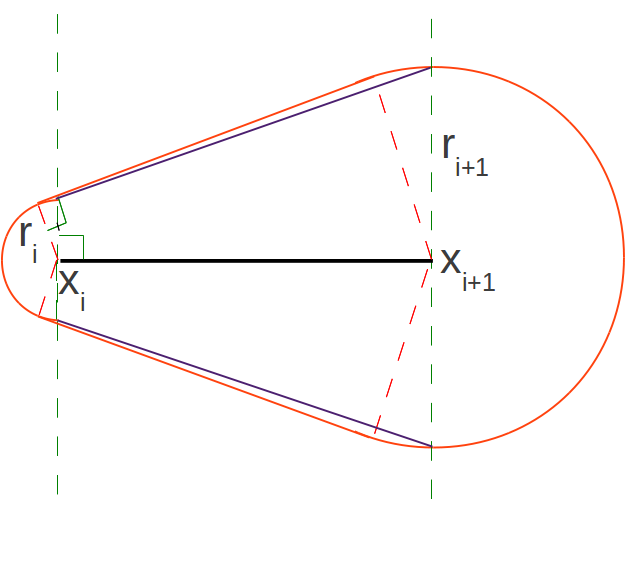}
\caption{The dilation $D_o$ obtained by using the orientation-based distance $d_o$ is represented by a full red line. The dilation $D_e$ obtained by using the simplified distance $d_e$ differs by $D_o$ only in the region delimited by the two dashed vertical lines, where it is represented by a blue line. However, the difference between the two dilations is very small also for large difference radii.}\label{dilschem}
\end{figure} 

%The signed distance of any point $y$ to the dilated objects is $D=d(y,\mathcal{S})-r(\lambda)$. Host Europe GmbH,
Denoting by $\mathcal{D}_o$ and $\mathcal{D}_e$ the dilations obtained by using the distances $d_o$ and $d_e$ respectively, as it can be observed  on the cartoon example shown in Fig. ~\ref{dilschem}: $d_o(\vec{y},\mathcal{S})=d_e(\vec{y},\mathcal{S})\sqrt{1-(\frac{\Delta r_i}{L_i})^2}-r_i(\lambda_e)$ in the region delimited by the vertical dashed green lines, and $D_o \approx D_e$ elsewhere. As it can be appreciated, the dilated models differ slightly even if the difference between the radii is large.  Since bacteria width has low variations, radii difference values are generally small compared to the segment length. Hence, the difference between the dilation models would be not significant.  Additionally, when using the simplified distance, the corresponding dilation contains less derivative terms and discontinuities, and consequently the variational optimization based on it would be computationally faster and more stable numerically. 
Hereafter, to simplify the notations, we will use the symbol $d$ to denote either $d_e$ or $d_o$, and $r$ to denote either $r_i(\lambda_e)$ or $r_i(\lambda_o)$ according to the desired distance model.

\paragraph{General remarks}

Although the distances $d_e$ and $d_o$ are continuous, both dilation outlines are made up of arc circles and straight lines. 
Note that the scale parameter of the outline is contained in the radius values: big or small outlines may be described with the same skeleton. Furthermore, some measures of interest inherent to the bacteria class are immediate with this representation: orientation, thickness, perimeter and length.

\subsection{Active skeletons}\label{snake}

\subsubsection{Skeleton initialization}
The optimization process relies on the availability of an initial skeleton for each bacterium. In this work, we derived it from the closed contours obtained by applying the bacteria segmentation method proposed in \cite{Primet2008Tracking}.
%In the discrete case it can be computed iteratively with simple morphological operators. One can filter this skeleton by removing the centers of small maximal disks. 
After computing the morphological skeleton through the use of morphological operators, a linear vectorization is performed by using an iterative method inspired to ~\cite{wall84,potier94} but based on angle variations. The vectorization starts from the farthest point from the gravity center of the skeleton. To take into account the presence of possible holes, a spline-interpolation is used to build a continuous skeletal line, which is finally sampled uniformly according to the wished accuracy of the skeleton.

\subsubsection{Skeleton optimization}

Given an image $u$ and the initial skeleton approximations $\lbrace S_i^0, i = 1,...,n\rbrace$, the aim of the active skeleton model is to  find the skeletons $\lbrace \hat{S}_i, i = 1,...,n \rbrace$ corresponding to the true medial axis of bacteria by evolving the initial skeletons. This is achieved by minimizing an energy function made up of two terms: a data-fidelity term and a regularity term that incorporates bacteria shape constraints. To handle the specific issue of several non-ovelapping objects, we propose an additional repulsion energy term. 

\begin{figure}[h]
\centering
\includegraphics[width=10cm]{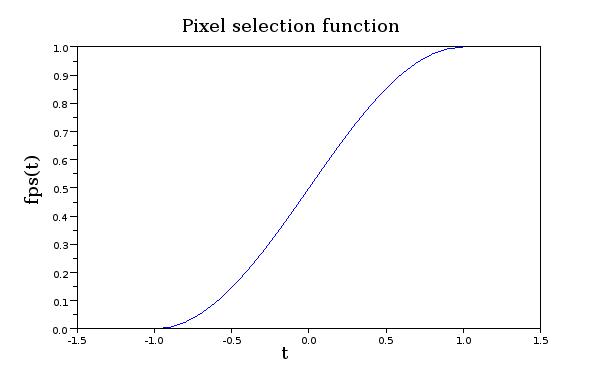}
\caption{Pixel selection function used to express the spatial condition. For pixels inside the closed contours the signed difference $r(x,\mathcal{S}) - d(x,\mathcal{S})$ has an absolute value less than 1, more than 1 otherwise. Therefore, the function being symmetric, we have that: $f_{ps}(d-r)+f_{ps}(r-d)=1$.}\label{psf}
\end{figure}

\paragraph{Data term}
Images captured by a phase-contrast microscopy are characterized by a difference of contrast between the foreground specimen and the background. Therefore, the data term has to be small when the difference of the contrast between foreground and background is large. 
Since time-lapse imaging objects contours may have large and slow variations (that is, strong low-frequency components), the use of a local neighborhood allows us to have a reasonable local model of the background. Typically, the data term corresponds to the difference between the average intensity inside the closed contour, and the average intensity in a ring surrounding it that corresponds to a high contrast between a bacterium and its local background. Since the objective function to be minimized has to be differentiable, the spatial condition is expressed through a smooth pixel selection function $f_{ps}$.
\begin{equation}
f_{ps}(t)=
\left\{\begin{array}{l}
\frac{1+\sin(\frac{\pi}{2}t)}{2} \quad t\in[-1,1]\\
0 \quad t<-1\\
1  \quad t>1
\end{array}\right.
\end{equation}

The data term $E_d$ is expressed as the difference between the internal energy $E_{in}$ and the external energy $E_{out}$, which are defined as follows
\begin{equation}
\left\{\begin{array}{l}
E_{in}= \displaystyle\sum\limits_{x \in I}\frac{f_{ps}[r(x,\mathcal{S}) - d(x,\mathcal{S})]f_c(x)}{f_{ps}[r(x,\mathcal{S}) - d(x,\mathcal{S})]} \\
E_{out}=\displaystyle\sum\limits_{x \in I | d(x,\mathcal{S})<r(x,\mathcal{S})+ \rho}\frac{f_{ps}[d(x,\mathcal{S}) - r(x,\mathcal{S})]f_c(x)}{f_{ps}[d(x,\mathcal{S}) - r(x,\mathcal{S})]}
\end{array}\right.
\end{equation}
where $f_c$ is an increasing smooth contrast function allowing to adapt the imaging contrast to the data, $\rho$ is the width of the ring surrounding the closed contours and, since $x \in s_i \subset \mathcal{S}$ can be written as $x= (1-\lambda) x_i + \lambda x_{i+1}$,  $r(x,\mathcal{S}) = (1-\lambda) r_i + \lambda r_{i+1}$.
It is easy to see that $f_{ps}[r(x,\mathcal{S}) - d(x,\mathcal{S})]$ selects pixels inside the closed contour, whereas $f_{ps}[d(x,\mathcal{S}) - r(x,\mathcal{S})]$ selects outside pixels.

\paragraph{Regularity terms}
Two regularity assumptions are embedded into the energy function by imposing a smooth segment angle variation and radii homogeneity. 

Most of the bacteria are weakly bent, therefore the skeleton curvature has to be as small as possible. The discrete curvature at point $x_i$ is measured by the angle of the arc at $x_i$, that corresponds to the angle  $\alpha_i = \widehat{\vec{x}_i-\vec{x}_{i-1},\vec{x}_{i+1}-\vec{x}_i}$.
Numerically, we compute $sin(\alpha_i)=\frac{det(\vec{x}_i-\vec{x}_{i-1},\vec{x}_{i+1}-\vec{x}_i)}{\Vert \vec{x}_i-\vec{x}_{i-1}\Vert \Vert \vec{x}_{i+1}-\vec{x}_i \Vert}$. The corresponding curvature-related regularization term is:
\begin{equation}
E_c=\sum\limits^{i=n-1}_{i=2}\sin^2(\alpha_i)
\end{equation}

Since the bacterium thickness is quite homogeneous, the difference between the radii of the skeleton points, $\{r_i, i = 1,...,n \}$, should be small.
%The term to term difference $r_{i+1}-r_{i}$ minimization has two drawbacks: it would give a proximity regularity highly dependent on the number of points constituting $S$ and moreover it would penalize a peanut-like shape. So we choose a global criterion instead. But of course the goal is not to move the radii artificially toward bigger or smaller values. Then the mean radius is not suitable because it can differ a lot according to some big or small extremity values. This is why the median radius value is prefered there. 
We consider the median radius value defined as $r_{med}=\underset{i\in[1..n]}{median}(r_i)$. The corresponding energy term is 
\begin{equation}
E_h=\sum\limits^{i=n}_{i=1}(r_i-r_{med})^2
\end{equation}

\paragraph{Repulsion term}

An additional repulsion term may be added when dealing with bacteria colonies. Denoting by $d(x_i,\mathcal{S}_k)$ the distance between a point $x_i \in \mathcal{S}_l$ and the skeleton $\mathcal{S}_k$, the corresponding distance between $x_i$ and the bacteria is $t=d(x_i,\mathcal{S}_k)-r_i-r^k_i$, where $r^k_i$ is the interpolated radius from $\mathcal{S}_k$ when computing the distance $d(x_i,\mathcal{S}_k)$. Since $t<0$ has to be avoided, by defining a repulsion function $f_{rep}(t)$ minimal when $t>0$, the repulsion energy term $E_r$ can be written as: 
\begin{equation}
E_r=\sum_k \sum_{x_i\in\mathcal{S}_l\neq\mathcal{S}_k} f_{rep}(d(x_i,S_l)-r_i-r^k_i)
\label{e4}
\end{equation}

\paragraph{Overall energy}

The global energy to minimize combines all above energy terms: $E=a E_d + b E_c + c E_h + d E_r$, where $a,b,c,d$ are positive weights. 
The energy minimization requires to compute derivatives according to points coordinates and radii. All derivatives are given in Appendix.

\section{Results}
\label{result}

The parameters defined in last section - $a$, $b$, $c$, $d$, $\rho$, the contrast function $f_c$ and the repulsion funcion $f_{rep}$ - were set once for all when correctly calibrated on our dataset. The experimental values were: $a=10$, $b=1$ , $c=0.01$, $d=0.1$, thickness $\rho=2$ pixels. The contrast function we used is $f_c(x)=\sin ^{0.8}I(x)$, which enhances the contrast between inside and outside mean values of bacteria. We have chosen  a quadratic repulsion function to penalize bigger overlaps: 
\begin{equation}
f_{rep}(t)=
\left\{\begin{array}{l}
(t-\Delta)^2 \text{ if } t<\Delta \\
0\text{ otherwise}
\end{array}\right.
\label{frep}
\end{equation}
with $\Delta=0.3$ pixels. 
%The post-dilation parameter $h$ was set to $1.5$ pixels when dealing with enlarged bacteria with the second method.
The algorithm was implemented in C with the MegaWave2 library \footnote{The software in C will be made publicly available with the publication of the article}. % with and without enlargement attempts defined in next paragraph~\ref{enlarge}. %Dilation models $\mathcal{D}_1$ and $\mathcal{D}_2$ were used with no significative differences. %Figure \ref{sksn1} shows an example.

\subsection{Method validation}
\label{validation}

%Delagado-Gonzalo et al. proposed a parametric snake, where the curve is described continuously by some coefficients using exponential B splines as  basis functions. Since the computational cost of spline snakes is determined in part by the size of the support of the basis function, they chosen the functions with the shortest support among all admissible functions.

% skeleton represents an object by a median line (the center line in the case of a tubular bacteria). Here we used one active skeleton for each cell, providing the long axis of the cell image (the median line) and its short axis (along the skeleton width). Active skeletons were adapted to bacteria in order to optimize the position of the skeleton in the image of the cell. 

\begin{figure}[h]
\centering
\includegraphics[width=9cm]{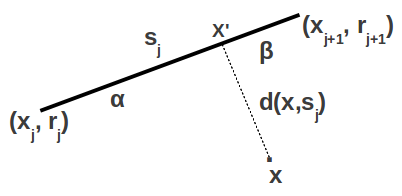}
\caption{The value $u(x)$ of the pixel $x$ is computed as a function of the distance of the pixel $x$ from the skeletal segment $s_j$ (dashed line).}
\label{dist2line}
\end{figure}

In this section, we compare our active skeleton model to the active contour model by using the Haussdorff distance to a synthetically generated ground truth on a set of synthetically generated images. Since both methods require a rough initialization we drew it by hand.
To generate the ground truth, as well as the test data, we started by considering a bacterial colony phase-contrast image. First, we computed the bacteria skeletons by applying the proposed method and then, by relying on the skeletons obtained, we recovered a synthetic image by computing each pixel value $u(x)$ as
\begin{equation}
u(x) = f \left(\inf_j\frac{d(x,s_j)}{r_j(x)} \right)
\end{equation}
where $d(x,s_j)$ is the distance of the pixel $x$ to the segment $s_j$ of a skeleton $S$ having as extremities the pixels $x_j$ and $x_{j+1}$ with associated radii $r_j$ and $r_{j+1}$ respectively (see Fig. \ref{dist2line}). $f$ is the Gauss error function and $r_j(x)$ is the interpolated radium computed on the projection of $x$ on $s_j$ ($x'$ in Fig. \ref{dist2line}). The interpolated radium value is computed as: $r_j(x) = \frac{\alpha r_j + \beta r_{j+1}}{\alpha + \beta}$ with $\alpha,\beta>0$.
 
\begin{figure}[h]
\centering
\includegraphics[width=7.2cm]{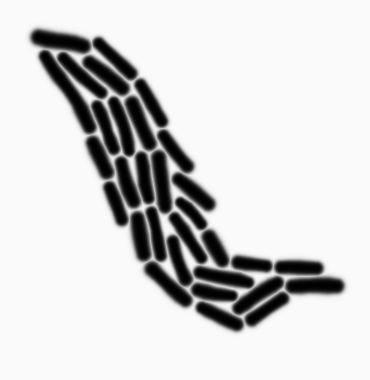} 
\caption{Synthetic image, whose noised versions constitute the test data used for validation.}\label{synthetic}
\end{figure}

\begin{figure}[h]
\centering
\includegraphics[width=11cm]{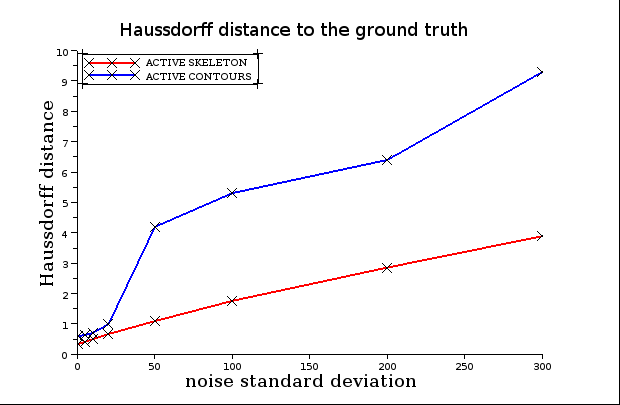}
\caption{Haussdorff distance of the active contour method (in blue) versus the Haussdorff distance of the active skeleton method (in red) as a function of the standard deviation of the Gaussian noise. As it can be observed, the active contour method is much more sensitive to the noise.}\label{SnakesvsSkelets}
\end{figure}

As test data, we used noised versions of the so generated synthetic image (see Figure \ref{synthetic}), obtained by adding a Gaussian white noise with increasing standard deviations. As ground truth, we used the boundaries from which the synthetic image has been generated. This is justified by the fact that, on the unnoised synthetic image, the average Haussdorff distances to ground truth of both methods are almost the same and of less than 1/10 of pixel (see Fig. \ref{SnakesvsSkelets}) for a standard deviation of noise equal to zero.

\begin{figure}[h]
\centering
\includegraphics[width=11cm]{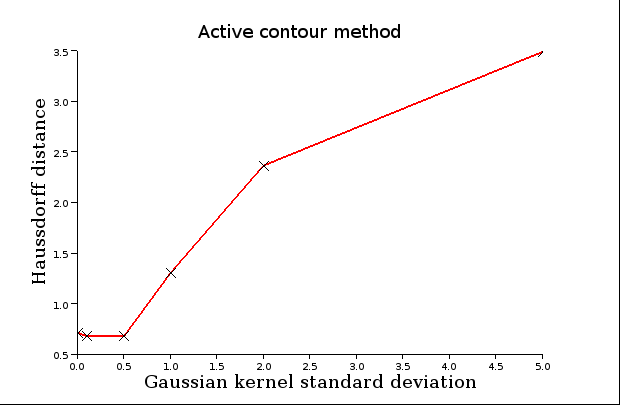}
\caption{Haussdorff distance to the ground truth of the active contour method for increasing levels of smoothness.}\label{smoothness}
\end{figure}

\begin{figure}[h]
\centering
\includegraphics[width=58mm]{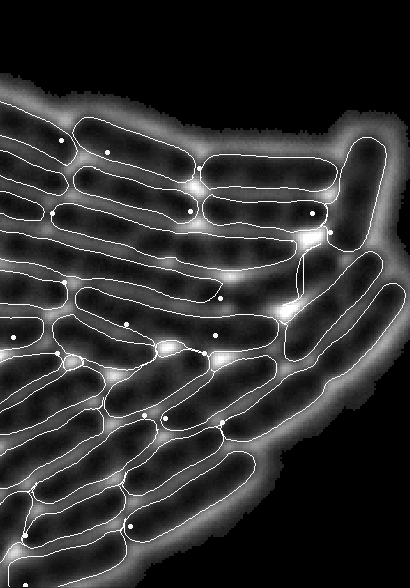}
\includegraphics[width=58mm]{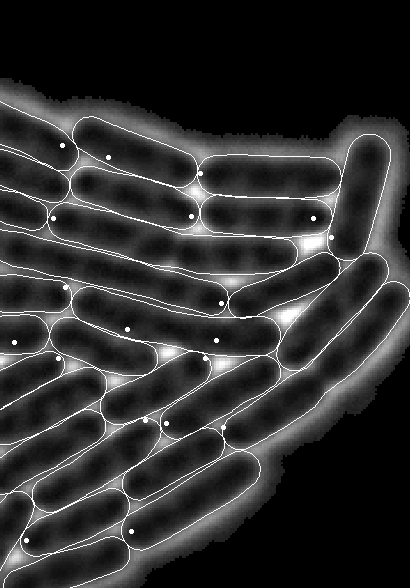}
\caption{(Left) Cells boundaries obtained by applying the snakes method. (Right) Cells boundaries obtained by applying the active skeleton method. Both used the method \cite{Primet2008Tracking} as initialization. Detected protein aggregates are visualized as white points. The accuracy of the active skeleton method allows a correct assignment of protein aggregates to cells.}\label{snakesVSskelets}
\end{figure}

Fig. \ref{SnakesvsSkelets} shows the Haussdorff distance to the ground truth of the active skeleton and the active contour methods. Since the active contour model assumes that the object contours are smooth, we previously smoothed the image by using a Gaussian kernel of standard deviation $\sigma_k = 0.5$ before applying this method. As it is shown in Fig.\ref{smoothness}, for relatively small values of the standard deviation of the Gaussian smoothing kernel, the performances of the active contour method slightly improves,  since the "edgness" of curves is enhanced by smoothing. However, there is an upper bound to the level of smoothing that can be applied which is related to the scale of the image structure we are looking for. The proposed active skeleton model allows to enforce smoothness without becoming excessively sensitive to noise by integrating an a priori on the shape which is itself smooth (rod or cigar-shape). It is worth to remark that on real images, the image contrast across cells boundaries is not constant as it is for our data test images. This greatly deteriorates the performances of the active contour method as it can be appreciated in Fig. \ref{snakesVSskelets}. This Fig. shows the interest of the proposed method in determining the association cell-aggregate. As detailed in section \ref{introduction}, solving this association requires accurate boundary estimation since fluorescent molecular components are usually concentrated at the poles of bacteria. 

\subsection{Application to the study of Escherichia Coli aging}

\begin{figure}[h]
\centering
\includegraphics[width=150mm]{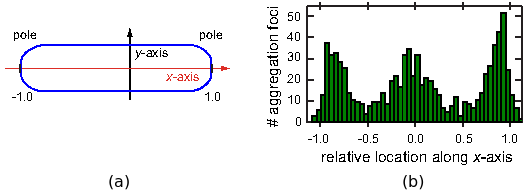}
\caption{(a) The $x-$ and $y-$ coordinates of  protein aggregate inside the cell correspond to the long and short axis respectively. (b) Histogram of the $x-$component of 1,644 images associated to initial trajectories. To take into account the high variability of cell lengths, the x-component was rescaled by division by the cell half-length, so that the cell poles are located at locations -1.0 and 1.0 respectively. Image adapted from the biological study  \cite{Coquel2013Localization}}\label{locationResults1}
\end{figure}

\begin{figure}[h]
\centering
\includegraphics[width=90mm]{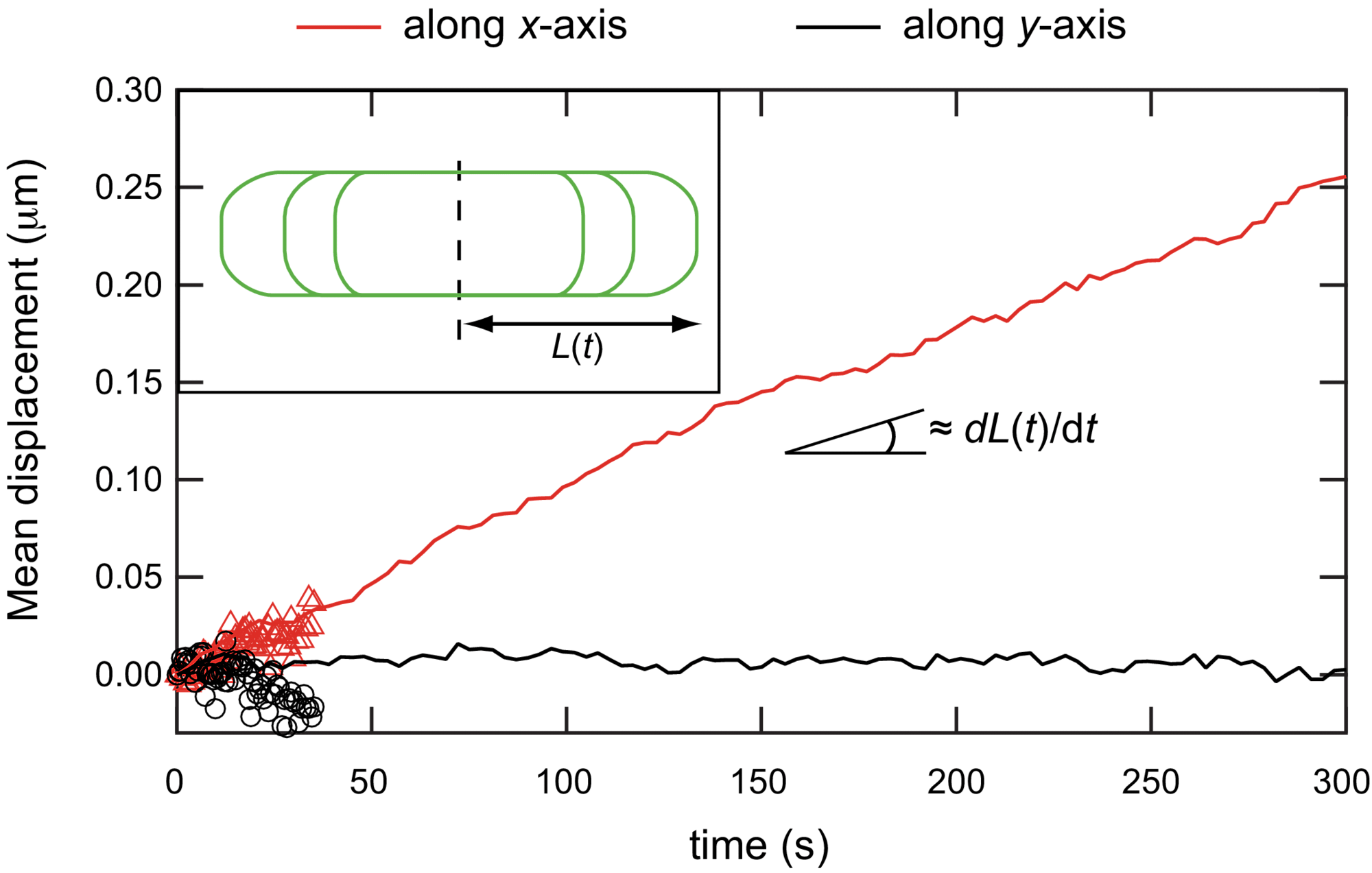}
\caption{Evolution mean displacement along the $x-$ (red) and $y-$ axis (black).    The movement along the x-axis  dominated the cell half-length growth.  Image adapted from the biological study \cite{Coquel2013Localization}.}\label{locationResults2}
\end{figure}

\begin{figure}[h!]
\centering
\begin{tabular}{ccrrrrrr}
\includegraphics[width=6cm]{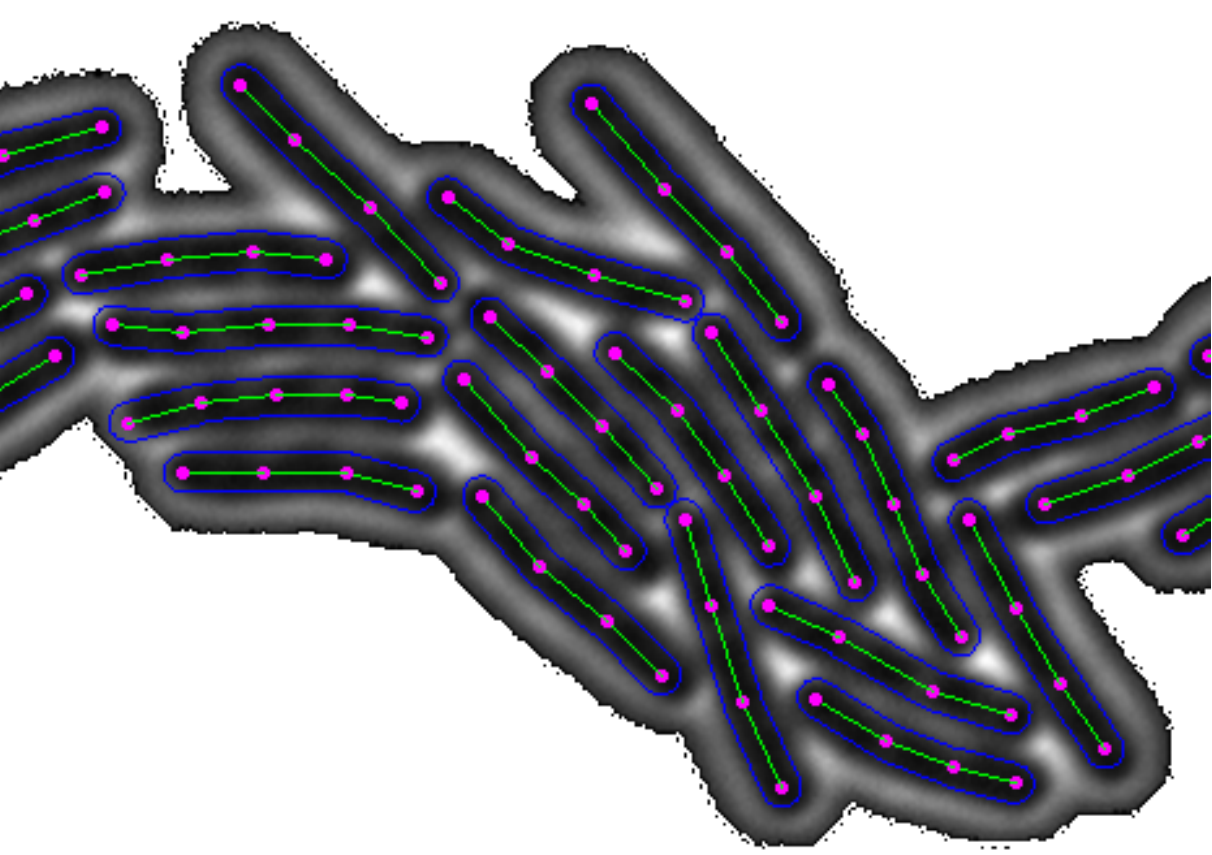}&
\includegraphics[width=6cm]{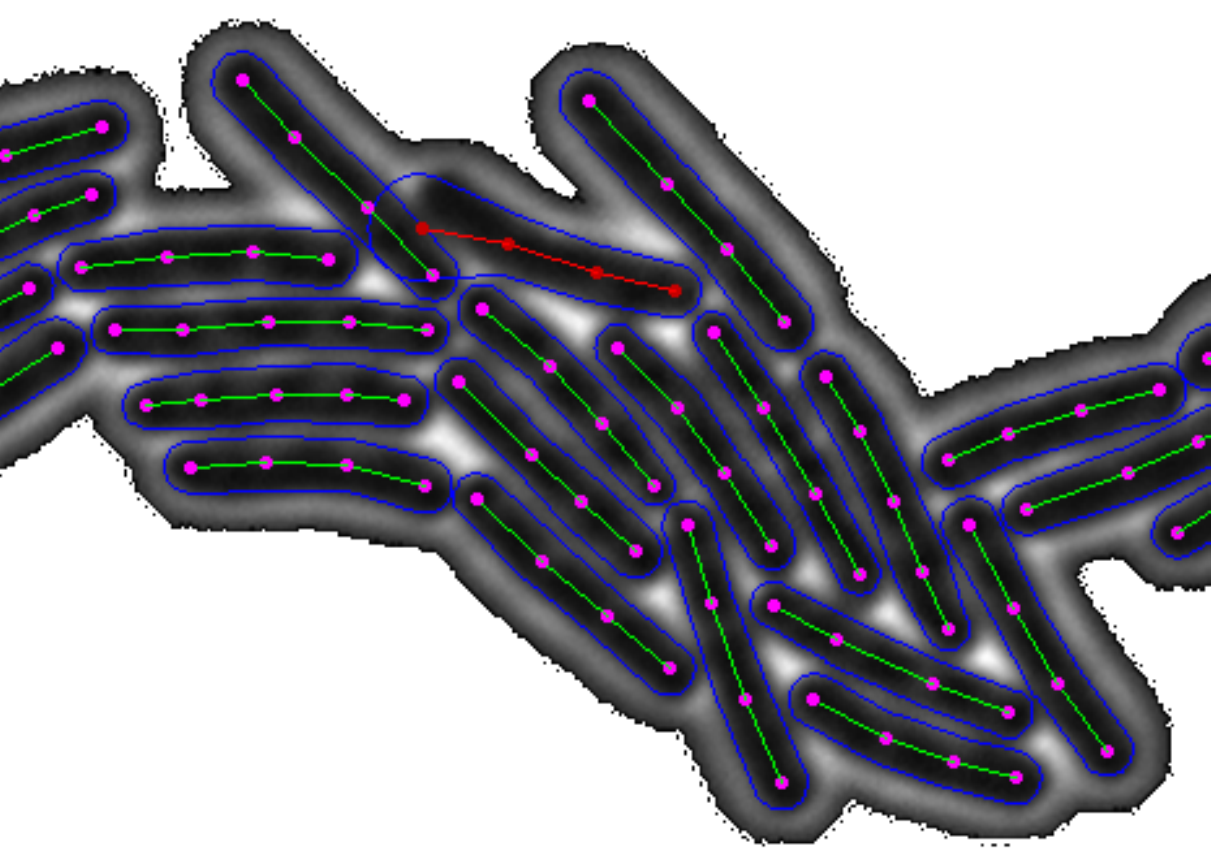}\\
(a) & (b)\\
\includegraphics[width=6cm]{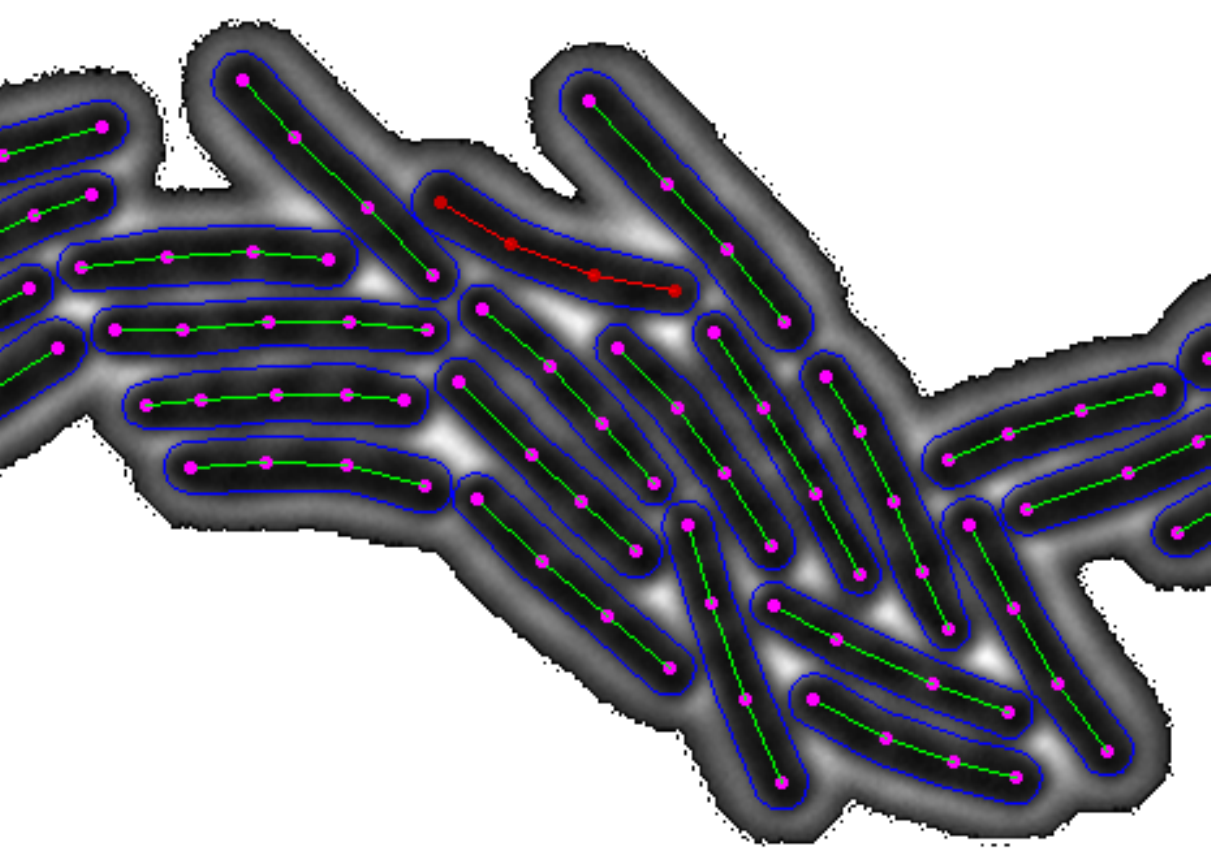}&
\includegraphics[width=6cm]{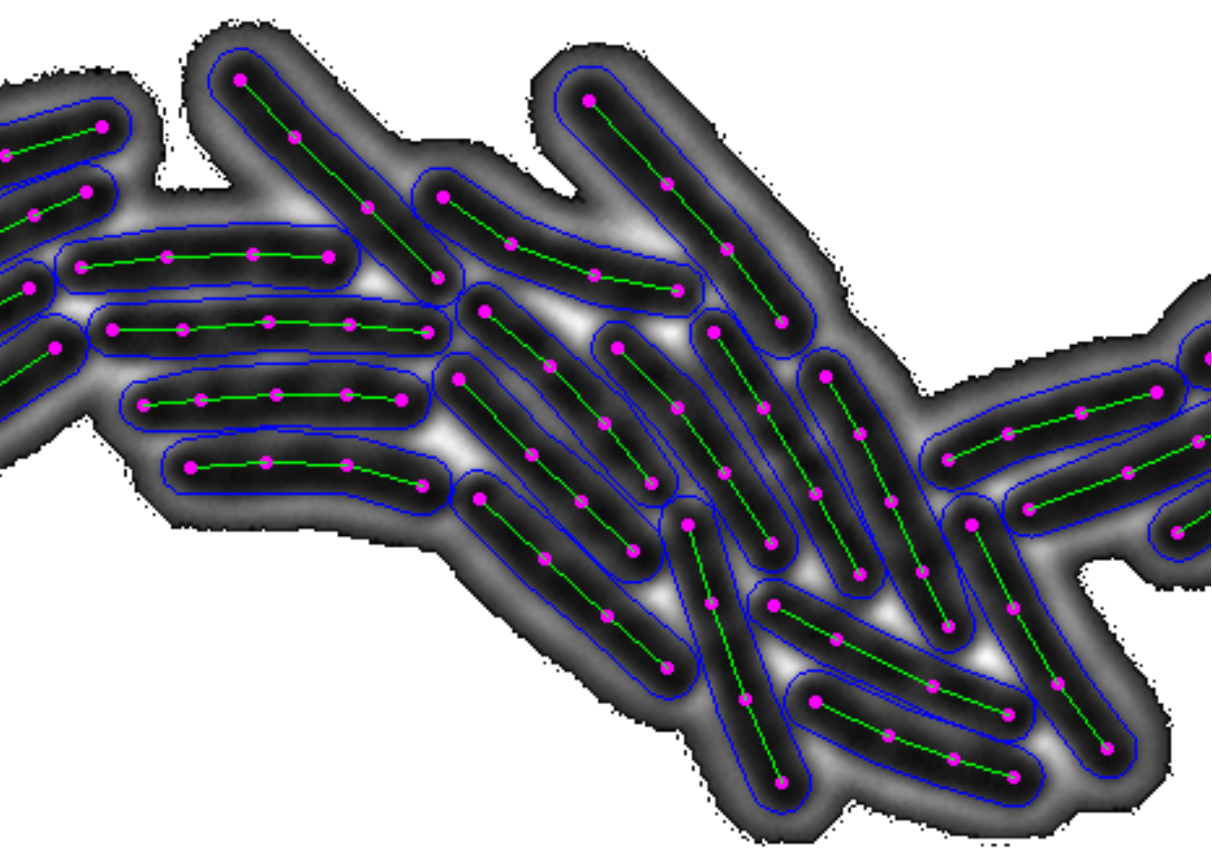}\\
(c) & (d)\\
\includegraphics[width=6cm]{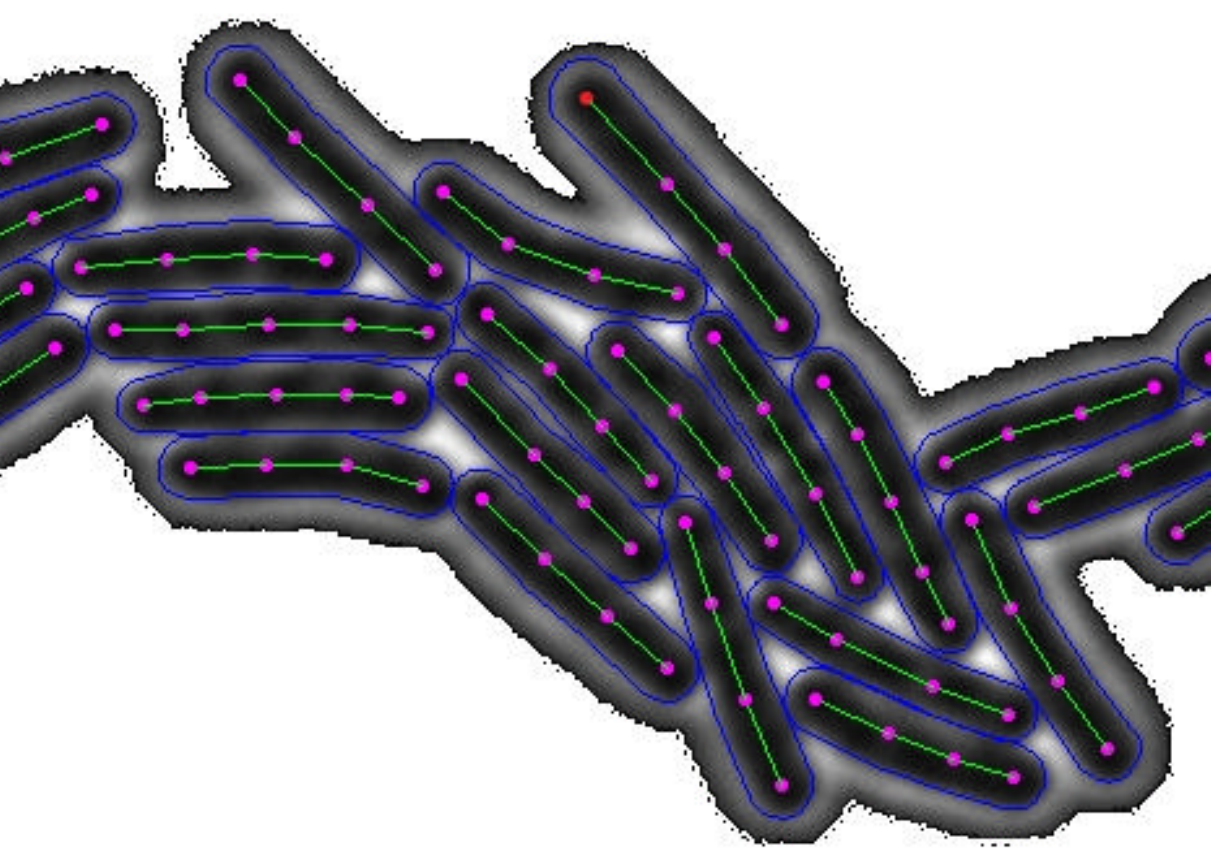}&
\includegraphics[width=6cm]{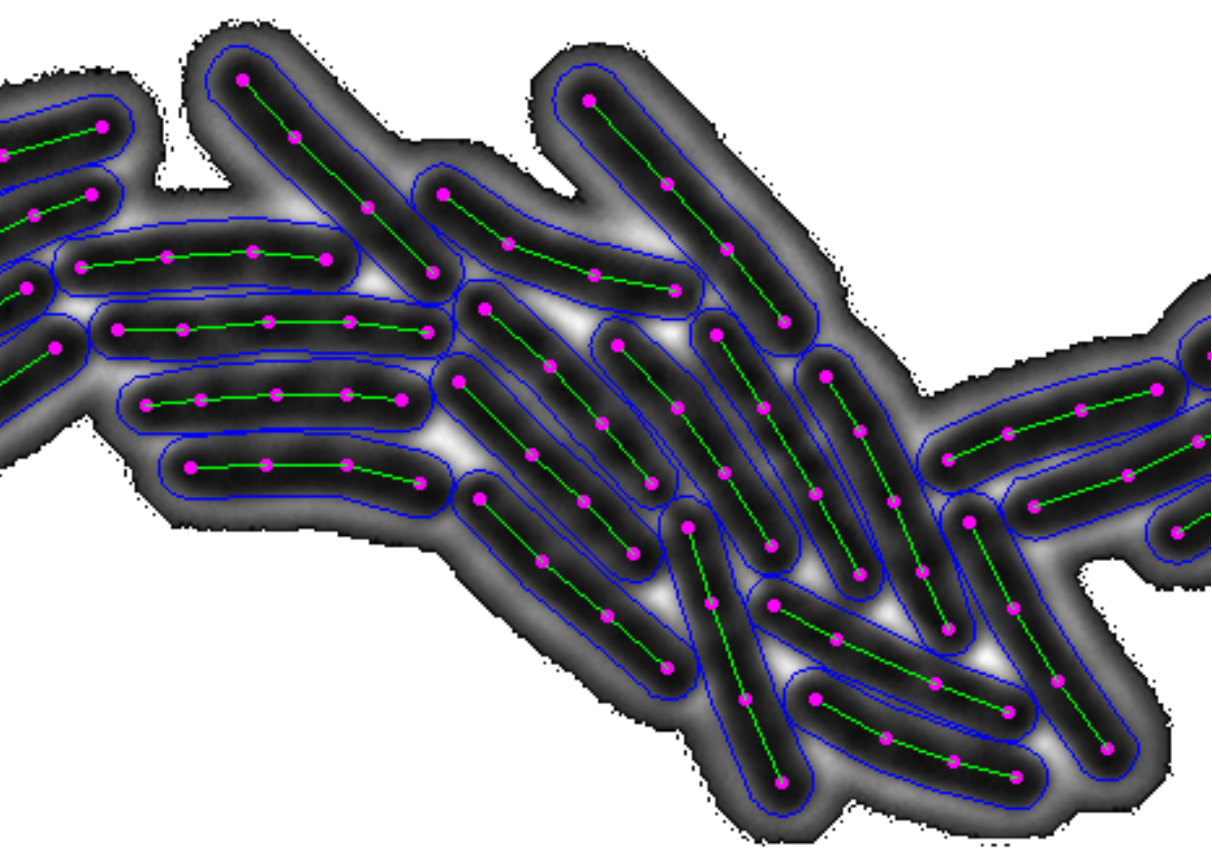}\\
(e)&(f)
\end{tabular}
\caption{(a) Initial state of active skeletons (green) made up of at least 4 points with the corresponding implicit outlines (blue). (b) Final state of active skeletons (green) and corresponding implicit outlines (blue) without the repulsion energy $E_r$. (c) Final state of active skeletons (green) and corresponding implicit outlines (blue)  with the repulsion energy $E_r$. The red skeleton was stopped by $E_r$. (d) Final state of active skeletons (green) and corresponding implicit outlines (blue) with a thickness parameter $\alpha_{in}=0.8$, still with the repulsion energy $E_r$. (e) Final state of active skeletons with a thickness parameter $\alpha_{in}=0.6$ and an additional bottom repulsion set to $\Delta_2=2$ pixels (effective on the red skeleton dots). (f) Final state of active skeletons with a repulsion of $\Delta+h=3.3$ pixels and a uniform dilation of $h=1.5$ pixels.}\label{ini}
\end{figure}

In Escherichia coli (E. coli) bacteria, aging-related protein aggregates accumulate at the old pole region of the aging bacterium. Studying the dynamics of these molecular components inside the cells requires automatic tools for 1) detecting protein aggregates in fluorescence images, 2) computing bacteria boundaries in the corresponding contrast-phase image, 3) assigning each protein aggregate to one cell, 4) expressing the coordinates of the protein aggregates in the basis composed  of the cell long (the median line) and short axes (along the skeleton width). We used the method in \cite{dimiccoli2015particle} to detect  protein aggregates in fluorescence images, the method \cite{Primet2008Tracking} for segmenting the cells in contrast-phase images, and the active skeleton method to refine bacteria boundaries prior to the affiliation of protein aggregates to cells. Then the final skeletons were used in two different manners. Firstly, to localize the protein aggregates in the skeletal short and long axes referential. Then, the simple shape of the skeleton was used to estimate the total cell width and length as that of the respective skeleton. 

As a convention, we refer below to the aggregate coordinate along the long axis as the x-coordinate and that along the short axis as the y-coordinate (see Fig. \ref{locationResults1} (a)). In Fig. \ref{locationResults1} (b) is shown the histogram of the $x-$component computed over 1,644 images. It can be seen that most aggregates accumulate along the center, that undergoes division, and the poles of bacteria. Fig. \ref{locationResults2}  schematizes the increase of the cell half-length during growth that dominates the movement along the x-axis, also showing the time evolution of the mean displacement.

\section{Discussion}

This study has covered a number of aspects of the problem of investigating the dynamics of bacteria cells and  their molecular components.  Overall the work reported produced numerical values for validation results and experimental results, and the discussion will be organised to highlight different parts of the process and to analyse the computational complexity. We will end this section by discussing the range of applicability of the proposed method.  An example of initialization is shown on figure~\ref{ini} (a): we choose a default number of four points per bacteria (and more if the spline interpolation is too far from the skeleton). Even if the morphological skeleton provides radii values at each skeleton point, we used a constant value corresponding to the minimal contact radius instead (see blue outlines). In this way, the influence of the initial segmentation obtained by applying the method proposed by Primet el al. \cite{Primet2008Tracking} is minimized. The results correspond to what is expected: the different final skeletons are inside the bacteria and the implicit outlines demarcate dark areas from light outsides. Figures \ref{ini} (b) and \ref{ini} (c) show final states from initial positions given in figure~\ref{ini} (a). They highlight the interest of the repulsion energy $E_r$. 
The second segmentation seems fully coherent with the initial image.

From a biological point of view the segmented bacteria are too thin since they should be in contact according to the experimental conditions. Of course this is not the case for the above segmentation (e.g on figure \ref{ini} (c)) because we consider that there are bright pixels outside: the bright pixels isolate the bacteria from each other. For this experimental reason, the bacteria have to be enlarged in some way. We propose two strategies to fix this problem. The first solution consists in adding a positive coefficient $\alpha_{in}<1$ in the expression $E_d=\alpha_{in}E_{in}-E_{out}$. This enables one to enlarge the bacteria according to the luminosity transitions as shown on figure~\ref{ini} (d). 
Though theoretically this is a good solution, the result is that different bacteria are more ore less dilated according to their outline intensity. Moreover, when taking low values of $\alpha_{in}$ (very big bacteria) the active skeleton may leak in the background. Another repulsion term from the bottom could be introduced but it supposes to set an additional arbitrary parameter: the minimal distance of bacteria to the bottom $\Delta_2$. Then the result is biased like on red dots in Fig. ~\ref{ini} (e). For the biological study we prefer another solution. We consider that bacteria are bigger than they appear on the image. Then the optimization is done by considering bigger bacteria in the repulsion energy term $E_r$ (interaction between real bacteria) while having a thin eroded representation of bacteria for the data confidence energy $E_d$. In other words, we assume that an eroded representation of the bacteria may fit the image. Let us call $h$ the radius difference between real and thin bacteria. To adapt consequently the repulsion term $E_r$, $\Delta$ in equation~(\ref{frep}) is replaced by $\Delta+h$ to prevent the overlapping of the physical bacteria. The optimization process produces thin bacteria. Then to obtain the physical bacteria from their thin image version, a homogeneous dilation is performed after the optimization. In practice, it simply consists in adding $h$ to all $r_i$ values. Figure~\ref{ini} (f) presents a $h=1.5$ pixels thickness difference.
%Since the real bacteria are bigger, we have to adapt consequently the repulsion term $E_4$ as follow: let us call $h$ the homogeneous dilation value. For the optimization process, $\Delta$ in equation~(\ref{frep}) is replaced by $\Delta+h$ to prevent the overlapping of the physical bacteria. Then the homogeneous dilation simply consists in adding $h$ to all $r_i$ values at the final state. Figure~\ref{ini} (f) presents a $h=1.5$ pixels postdilation.
% over the above solution is that we control the dilation explicitly: $2.h$ is the exact thickness increase when enough space between bacteria is available. 

The computational complexity of the proposed method depends on three parameters: the number of active skeletons, the number of skeletal points used to model each of them and the image resolution. 
More precisely, for one step of the gradient descent and for a single skeleton, the complexity of deriving the energy term is $\mathcal{O}(MN)$, where $N$ is the number of skeleton points and $M$ is the number of pixels within a box surrounding the skeleton dilation. For $K$ skeletons without repulsion energy, the complexity is $\mathcal{O}(KMN)$. When considering the repulsion energy, it becomes $\mathcal{O}(KMN) + O(K^2N^2)$, where the second term corresponds to inter-bacteria distances. Since accurate segmentation of bacteria can be achieved  with a few skeletal points,  the algorithm complexity will mainly depend on the image resolution and on the number of bacteria in this particular application.

We conclude by remarking that although this work was tailored to cigar- or rod-shaped bacteria, it would work as well for any kind of curved ovoid shape. For non-ovoid shapes, it could be adapted by changing the dilation model. For more complicated ramified shapes, the skeleton model itself should be adapted. The utility spectra of the proposed method is not restricted to bacteria, but to all cases in which a precise segmentation of cigar- or rod-shaped object is needed and the image quality is poor. For instance it could be very useful for segmenting images of insects in the study of insect populations, or for the segmentation and recognition of tumors and cigar- or rod-shaped cells.

\section{Conclusions}
This paper has proposed an active skeleton approach for bacteria modeling, that presents several advantages for the study of bacteria shape properties and the dynamics of their molecular components. Indeed, the long and short axis of the skeleton define a reference system centered to the bacterium, which is crucial to study the dynamics of subcellular components inside the cell. In addition, the computation of shape properties such as local orientation, thickness, length and perimeter from the skeleton representation is straightforward. 
Finally, bacteria boundaries are computed accurately even in very noised images by introducing implicitly smooth shape priors into the segmentation process. The improved accuracy is important to reduce the ambiguity of the association molecule/cell in time-lapse fluorescence and phase-contrast imaging.
%The class of shapes corresponding to the model is very restricted compared to snakes but offers a greater robustness when adapted to the target. 
%A bacteria segmentation has been presented with a region-based approach and thickness enhancement methods. 

The proposed model has been successfully used to study the dynamic of protein aggregates in \textit{E. Coli} \cite{Coquel2013Localization}. Localization of proteins to specific positions inside bacteria is crucial to several physiological processes, including chromosome organization, chemotaxis or cell division and aging. The Image-J plugin of the proposed method can be found online at \textit{http://fluobactracker.inrialpes.fr}.

Further improvements could be discussed and developed according to the kind of data and to the imaging task to deal with. For example, the fidelity term of the energy functional could be chosen differently from the literature on data energy terms for snakes. Ad-hoc initializations could be derived from morphological gray-level skeletons and segment extraction. Finally, the skeleton model could handle more complex shapes, with more branches and angle constraints, and eventually with other distance definitions relaxing the circular extremity of each branch.

 \section*{Acknowledgement} This work has been partially supported by the project PAGDEG: Causes and consequences of protein aggregation in cellular degeneration. Funding: French ANR. The authors would like to thank Anne-Sophie Coquel for sharing the biological data and contributing to validate the proposed method.

\bibliographystyle{gCMB}
\bibliography{mybibfile}
\end{document}

\begin{frontmatter}

\title{Appendix}

\end{frontmatter}

%%%%%%%%%%%%%%%%%%%%%%%%%%%%%%%%%%%%
The distance of a point $y \in \Omega$ to a skeleton $\mathcal{S}$ is defined as the minimum of distances of the point to each segment of the skeleton.
$d(\vec{y},\mathcal{S})=\displaystyle{\min_{i = 1,...,n-1}}\Big \lbrace d(\vec{y},s_i) \Big \rbrace$. This implies that $\exists i\in[1..n-1]$ such that 
$d(\vec{y},\mathbf{S})=d(\vec{y},s_i)$. Then when $\vec{y}$ and $\mathcal{S}$ are settled, $d(\vec{y},\mathcal{S})$ only depends on $\vec{x}_i$ and $\vec{x}_{i+1}$. It means that $\frac{\partial d(\vec{y},\mathbf{S})}{\partial x_j}\not=0$ if and only if $j=i$ or $j=i+1$.

Then the calculus of the derivatives of $d(\vec{y},\mathcal{S})$ is restricted to the calculus of $\frac{\partial d(\vec{y},s_i)}{\partial x_i}$, $\frac{\partial d(\vec{y},s_i)}{\partial x_{i+1}}$, $\frac{\partial d(\vec{y},s_i)}{\partial y_i}$, $\frac{\partial d(\vec{y},s_i)}{\partial y_{i+1}}$, and $\frac{\partial d(\vec{y},s_i)}{\partial r_i}$, $\frac{\partial d(\vec{y},s_i)}{\partial r_{i+1}}$. We give them for $x_i$, $x_{i+1}$, $r_i$ and $r_{i+1}$. The results for $y_i$ and $y_{i+1}$ are obtained similarly.

\section{Partial derivatives of the energy data term $E_d$}

\subsection{Simplified dilation}
Let us consider $\lambda_e$, $d_e$ and $\tilde{r}_1=(1-\lambda_e)r_i+\lambda_e r_{i+1}=r_i+\lambda_e\Delta r_i$. The desired derivative for the data energy term $E_d$ is that of $d_e-\tilde{r}_1$.

Derivatives for $0<\lambda_1<1$ according to $x_i$, $x_{i+1}$ :
\begin{equation}
\frac{\partial \lambda_1}{\partial x_i}=\frac{x_i-x+(2\lambda_1-1)\Delta x_i}{L_i^2} \text{ and }
\frac{\partial \lambda_1}{\partial x_{i+1}}=\frac{x-x_i+(2\lambda_1-1)\Delta x_i}{L_i^2}
\end{equation}
\begin{equation}
\frac{\partial d_e}{\partial x_i}=\frac{(1-\lambda_1)(x_i-x+\lambda_1\Delta x_i)}{d_e} \text{ and }
\frac{\partial d_e}{\partial x_{i+1}}=\frac{\lambda_1(x_i-x+\lambda_1\Delta x_i)}{d_e}
\end{equation}
\begin{equation}
\frac{\partial \tilde{r}_1}{\partial x_i}=\frac{\partial \lambda_1}{\partial x_i}\Delta r_i \text{ and }
\frac{\partial \tilde{r}_1}{\partial x_{i+1}}=\frac{\partial \lambda_1}{\partial x_{i+1}}\Delta r_i
\end{equation}
So that all together :

\begin{equation}
\left\{\begin{array}{l}
\frac{\partial (d_e-\tilde{r}_1)}{\partial x_i}=(\frac{(1-\lambda_1)}{d_e}-\frac{\Delta r_i}{L_i^2})(x_i-x)+(\frac{\lambda_1(1-\lambda_1)}{d_e}-\frac{(2\lambda_1-1)\Delta r_i}{L_i^2})\Delta x_i
\text{ and }\\
\frac{\partial (d_e-\tilde{r}_1)}{\partial x_{i+1}}=(\frac{\lambda_1}{d_e}+\frac{\Delta r_i}{L_i^2})(x_i-x)+(\frac{\lambda_1^2}{d_e}+\frac{2\lambda_1\Delta r_i}{L_i^2})\Delta x_i
\end{array}\right.
\label{d1x}
\end{equation}

Derivatives for $0<\lambda_1<1$ according to $r_i$, $r_{i+1}$ :\\
$d_e$ does not depend on $r_i$ nor $r_{i+1}$.
\begin{equation}
\left\{\begin{array}{l}
\frac{\partial (d_1-\tilde{r}_1)}{\partial r_i}=-\frac{\partial \tilde{r}_1}{\partial r_i}=\lambda_1-1
\text{ and }\\
\frac{\partial (d_1-\tilde{r}_1)}{\partial r_{i+1}}=-\frac{\partial \tilde{r}_1}{\partial r_{i+1}}=-\lambda_1
\end{array}\right.
\label{d1r}
\end{equation}

Equations \ref{d1x} and \ref{d1r} give the final formula to compute for $E_d$ (data energy) derivates in the case of approximate dilation calculus.

\subsection{Oriented dilation}
Now we work with $\lambda_2 =\lambda_1+\lambda$, $d_o$ and $\tilde{r}_2=(1-\lambda_o)r_i+\lambda_o r_{i+1}=r_i+\lambda_o\Delta r_i$. The desired derivative for the data energy term $E_1$ is that of $d_o-\tilde{r}_2$.

The derivatives for $0<\lambda_1+\lambda <1$ with respect to $x_i$ and $x_{i+1}$ are as follows:
\begin{equation}
\frac{\partial \lambda }{\partial x_i}=\frac{\lambda}{d_e^2}[(1-\lambda_1)(x_i-x+\lambda_1\Delta x_i)+\frac{d_e^2+d_o^2}{Li_2}\Delta x_i]
 \text{ and }
\frac{\partial \lambda }{\partial x_{i+1}}=\frac{\lambda}{d_e^2}[\lambda_1(x_i-x+\lambda_1\Delta x_i)-\frac{d_e^2+d_o^2}{Li_2}\Delta x_i]
\end{equation}
\begin{equation}
\frac{\partial d_o}{\partial x_i}=\frac{d_o}{d_e^2}[(1-\lambda_1)(x_i-x+\lambda_1\Delta x_i)+\lambda ^2\Delta x_i]
 \text{ and }
\frac{\partial d_o}{\partial x_{i+1}}=\frac{d_o}{d_e^2}[\lambda_1(x_i-x+\lambda_1\Delta x_i)-\lambda ^2\Delta x_i]
\label{d2dx}
\end{equation}
\begin{equation}
\left\{\begin{array}{l}
\frac{\partial \tilde{r}_2}{\partial x_i}=[(\frac{1}{L_i^2}+\frac{\lambda(1-\lambda_1)}{d_e^2})(x_i-x)+(\frac{2\lambda_1-1}{L_i^2}+\frac{\lambda_1\lambda(1-\lambda_1)}{d_e^2}+\frac{\lambda(d_e^2+d_o^2)}{d_e^2L_i^2})\Delta x_i]\Delta r_i
\text{ and }\\
\frac{\partial \tilde{r}_2}{\partial x_{i+1}}=[(-\frac{1}{L_i^2}+\frac{\lambda_1\lambda }{d_e^2})(x_i-x)+(\frac{-2\lambda_1 }{L_i^2}+\frac{\lambda_1^2\lambda }{d_e^2}-\frac{\lambda(d_e^2+d_o^2)}{d_e^2L_i^2})\Delta x_i]\Delta r_i
\end{array}\right.
\label{d2rx}
\end{equation}
From the previous equations one can deduce $\frac{\partial (d_o-\tilde{r}_2)}{\partial x_i}$ and $\frac{\partial (d_o-\tilde{r}_2)}{\partial x_{i+1}}$.

Derivatives for $0<\lambda_1+\lambda <1$ according to $r_i$, $r_{i+1}$:\\
Note that here $d_o$ (and $\lambda$) depends on $r_i$ and $r_{i+1}$.
\begin{equation}
\frac{\partial \lambda }{\partial r_i}=-\frac{d_o}{L_i^2-\Delta r_i^2}=-\frac{\lambda L_i^2}{\Delta r_i(L_i^2-\Delta r_i^2)}
 \text{ and }
\frac{\partial\lambda }{\partial r_{i+1}}=\frac{d_2}{L_i^2-\Delta r_i^2}=\frac{\lambda L_i^2}{\Delta r_i(L_i^2-\Delta r_i^2)}
\end{equation}
\begin{equation}
\frac{\partial d_2}{\partial r_i}=-\frac{d_2^2}{d_1^2}\lambda \Delta r_i^2
 \text{ and }
\frac{\partial d_2}{\partial r_{i+1}}=\frac{d_2^2}{d_1^2}\lambda \Delta r_i^2
\end{equation}
\begin{equation}
\frac{\partial \tilde{r}_2}{\partial r_i}=1-\lambda_1-\lambda \frac{d_1^2+d_2^2}{d_1^2}
 \text{ and }
\frac{\partial \tilde{r}_2}{\partial r_{i+1}}=\lambda_1+\lambda \frac{d_1^2+d_2^2}{d_1^2}
\end{equation}
\begin{equation}
\frac{\partial (d_2-\tilde{r}_2)}{\partial r_i}=-1+\lambda_1+\lambda \frac{d_1^2+d_2^2(1-\Delta r_i^2)}{d_1^2}
 \text{ and }
\frac{\partial (d_2-\tilde{r}_2)}{\partial r_{i+1}}=-\lambda_1-\lambda \frac{d_1^2+d_2^2(1-\Delta r_i^2)}{d_1^2}
\label{d2r}
\end{equation}
Equations \ref{d2dx}, \ref{d2rx} and \ref{d2r} give access to the derivatives of $E_d$ (data energy) derivates in the case of real dilation calculus.

\section{Other derivatives}

\subsection{Derivation of the energy term $E_c$ (curvature)}
Each curvature term $\sin\alpha_i=\frac{det(\overrightarrow{\vec{x}_{i-1}\vec{x}_i},\overrightarrow{\vec{x}_i\vec{x}_{i+1}})}{\Vert\overrightarrow{\vec{x}_{i-1}\vec{x}_i}\Vert \Vert\overrightarrow{\vec{x}_i\vec{x}_{i+1}}\Vert}$ has to be derived. 
$$\sin\alpha_i=\frac{(x_i-x_{i-1})(y_{i+1}-y_i)-(x_{i+1}-x_i)(y_i-y_{i-1})}{\sqrt{(\Delta x_{i-1}^2+\Delta y_{i-1}^2)(\Delta x_i^2+\Delta y_i^2)}}=\frac{A_i}{B_i}$$
Of course $\frac{\partial (\sin\alpha_i)}{\partial x_j}=0$ if $j\not=i$, $j\not=i-1$ and $j\not=i+1$. 
The partial derivatives of the numerator are:
$$
\left\{\begin{array}{l}
\frac{\partial A_i}{\partial x_i}=y_{i+1}-y_{i-1}\\
\frac{\partial A_i}{\partial x_{i-1}}=-(y_{i+1}-y_i)\\
\frac{\partial A_i}{\partial x_{i+1}}=-(y_i-y_{i-1})
\end{array}\right.
$$
The partial derivatives of the denominator are:
$$\frac{\partial(B_i^2)}{\partial x_i}=2\Delta x_{i-1}(\Delta x_i^2 + \Delta y_i^2)  + 2\Delta x_i(\Delta x_{i-1}^2 + \Delta y_{i-1}^2)=2B_i\frac{\partial B_i}{\partial x_i}$$
Analogously:
$$
\left\{\begin{array}{l}
\frac{\partial B_i}{\partial x_i}=\frac{\Delta x_{i-1}(\Delta x_i + \Delta y_i)^2(\Delta_i)^2-\Delta x_i(\Delta x_{i-1}^2 + \Delta y_{i-1}^2))}{B_i}\\
\frac{\partial B_i}{\partial x_{i-1}}=-\frac{\Delta x_{i-1}(\Delta x_i^2 + \Delta y_i^2) }{B_i}\\
\frac{\partial B_i}{\partial x_{i+1}}=\frac{\Delta x_i(\Delta x_{i-1}^2 + \Delta y_{i-1}^2)}{B_i}
\end{array}\right.
$$

Then finally one has~:
\begin{equation}
\left\{\begin{array}{l}
\frac{\partial \sin\alpha_i}{\partial x_i}=\frac{y_{i+1}-y_{i-1}}{B_i}-\frac{A_i}{B_i^3}((\Delta x_{i-1}(\Delta x_i^2 + \Delta y_i^2)-\Delta x_i(\Delta x_{i-1}^2 + \Delta y_{i-1}^2)))\\
\frac{\partial \sin\alpha_i}{\partial x_{i-1}}=\frac{-(y_{i+1}-y_i)}{B_i}+\frac{A_i}{B_i^3}(\Delta x_{i-1}(\Delta x_i^2 + \Delta y_i^2))\\
\frac{\partial \sin\alpha_i}{\partial x_{i+1}}=\frac{-(y_i-y_{i-1})}{B_i}-\frac{A_i}{B_i^3}(\Delta x_i(\Delta x_{i-1}^2 + \Delta y_{i-1}^2))
\end{array}\right.
\end{equation}
and symmetrically
\begin{equation}
\left\{\begin{array}{l}
\frac{\partial \sin\alpha_i}{\partial y_i}=\frac{x_{i+1}-x_{i-1}}{B_i}-\frac{A_i}{B_i^3}((\Delta y_{i-1}(\Delta x_i^2 + \Delta y_i^2)-\Delta y_i(\Delta x_{i-1}^2 + \Delta y_{i-1}^2)))\\
\frac{\partial \sin\alpha_i}{\partial y_{i-1}}=\frac{-(x_{i+1}-x_i)}{B_i}+\frac{A_i}{B_i^3}(\Delta y_{i-1}(\Delta x_i^2 + \Delta y_i^2))\\
\frac{\partial \sin\alpha_i}{\partial y_{i+1}}=\frac{-(x_i-x_{i-1})}{B_i}-\frac{A_i}{B_i^3}(\Delta y_i(\Delta x_{i-1}^2 + \Delta y_{i-1}^2))
\end{array}\right.
\end{equation}

Then the derivation of $\sin^2\alpha_i$ is immediate~: 
$\frac{\partial \sin^2\alpha_i}{\partial .}=2\sin\alpha_i \frac{\partial \sin\alpha_i}{\partial .}=2\frac{A_i}{B_i}\frac{\partial \sin\alpha_i}{\partial .}$. And when $i\in[1..n]$ is set\footnote{for $i=0$ and $i=n$ we note $\sin\alpha_1=\sin\alpha_n=0$}~:
\begin{equation}
\left\{\begin{array}{l}
\frac{\partial E_2}{\partial x_i}=\frac{\partial}{\partial x_i}(\sin^2\alpha_{i-1}+\sin^2\alpha_i+\sin^2\alpha_{i+1})\\
\frac{\partial E_2}{\partial y_i}=\frac{\partial}{\partial y_i}(\sin^2\alpha_{i-1}+\sin^2\alpha_i+\sin^2\alpha_{i+1})
\end{array}\right.
\label{eqE2}
\end{equation}
 
\subsection{Derivation of the energy term $E_h$ (radius homogeneity)}
Assuming that $r_{med}$ does not depend on $r_i$, the partial derivative of the energy $E_h$ with respect to the variable $r_i$ is as follows:

\begin{equation}
\frac{\partial E_h}{\partial r_i}=2(r_i-r_{med})
\label{der:eqE3}
\end{equation}

\subsection{Derivation of the energy term $E_r$ (repulsion)}
To compute the partial derivatives of the energy $E_r$ with respect to the variables $x_i$, $y_i$, and $r_i$ of the skeleton $\mathcal{S}_k$, only the $k^{th}$ term of $E_r$ is taken into account.
$\left\{\begin{array}{lll}
\frac{\partial E_r}{\partial x_i}=\displaystyle{\sum_{i,l\neq k} \frac{\partial d(\vec{x}_i,\mathcal{S}_l)}{\partial x_i} f_{rep}'(t) }=\displaystyle{\sum_{i,l\neq k} \frac{x_i-x_k^l}{d(\vec{x}_i,S_l)} f_{rep}'(t) }\\
\frac{\partial E_r}{\partial y_i}=\displaystyle{\sum_{i,l\neq k} \frac{\partial d(\vec{x}_i,\mathcal{S}_l)}{\partial y_i} f_{rep}'(t) }=\displaystyle{\sum_{i,l\neq k} \frac{y_i-y_k^l}{d(\vec{x}_i,S_l)} f_{rep}'(t) }\\
\frac{\partial E_r}{\partial r_i}=\displaystyle{\sum_{i,l\neq k} \frac{\partial d(\vec{x}_i,\mathcal{S}_l)}{\partial r_i} f_{rep}'(t) }=\displaystyle{\sum_{i,l\neq k} - f_{rep}'(t) }
\end{array}\right.\label{der:eqE4}$